\documentclass[lettersize,journal]{IEEEtran}
\usepackage{amsmath,amsfonts}
\usepackage{algorithmic}
\usepackage{algorithm}
\usepackage{array}
\usepackage{booktabs}
\usepackage[caption=false,font=footnotesize,labelfont=rm,textfont=rm]{subfig}
\usepackage{textcomp}
\usepackage{stfloats}
\usepackage{url}
\usepackage{verbatim}
\usepackage{graphicx}
\usepackage{cite}
\usepackage{float}  
\usepackage{multirow}
\usepackage{hyperref} 
\usepackage{utfsym} 
\begin{document}

\title{A Hierarchical Multi-Vehicle Coordinated Motion Planning Method based on Interactive Spatio-Temporal Corridors}

\author{Xiang Zhang, Boyang Wang, Yaomin Lu, Haiou Liu, Jianwei Gong, and Huiyan Chen
\thanks{This work was supported in party by the National Natural Science Foundation of China under Grant 52172378 and in part by the Beijing Institute of Technology Fund Program for Young Scholars. (Corresponding author: Boyang Wang)}
\thanks{Xiang Zhang, Boyang Wang, Yaomin Lu, Haiou Liu, Jianwei Gong, and Huiyan Chen are with the School of Machianical Engineering, Beijing Institute of Technology, Beijing 100081, China (e-mail: xiangzhang0825@163.com, boyang\_wang@bit.edu.cn, luyaomin9612@163.com, bit\_lho@bit.edu.cn, gongjianwei@bit.edu.cn, Chen\_h\_y@263.net)}}

\markboth{IEEE Transactions on Intelligent Vehicles, Vol.XXX, No.XXX, March 2023}%
{Shell \MakeLowercase{\textit{et al.}}: A Sample Article Using IEEEtran.cls for IEEE Journals}


\maketitle

\begin{abstract}
Multi-vehicle coordinated motion planning has always been challenged to safely and efficiently resolve conflicts under non-holonomic dynamic constraints. Constructing spatial-temporal corridors for multi-vehicle can decouple the high-dimensional conflicts and further reduce the difficulty of obtaining feasible trajectories. Therefore, this paper proposes a novel hierarchical method based on interactive spatio-temporal corridors (ISTCs). In the first layer, based on the initial guidance trajectories, Mixed Integer Quadratic Programming is designed to construct ISTCs capable of resolving conflicts in generic multi-vehicle scenarios. And then in the second layer, Non-Linear Programming is settled to generate in-corridor trajectories that satisfy the vehicle dynamics. By introducing ISTCs, the multi-vehicle coordinated motion planning problem is able to be decoupled into single-vehicle trajectory optimization problems, which greatly decentralizes the computational pressure and has great potential for real-world applications. Besides, the proposed method searches for feasible solutions in the 3-D $(x,y,t)$ configuration space, preserving more possibilities than the traditional velocity-path decoupling method. Simulated experiments in unsignalized intersection and challenging dense scenarios have been conduced to verify the feasibility and adaptability of the proposed framework.
\end{abstract}


\section{Introduction}
\IEEEPARstart{M}{ulti-vehicle} collaboration technology is attracting more and more attentions due to its potential value in traffic, port, warehouse and other scenarios. The rapid development of vehicle-to-everything (V2X) technology has made it possible for vehicles to interact with information from surrounding vehicles, infrastructures and command terminals \cite{siegel2017survey}. On this basis, the multi-vehicle planning problem can no longer be solved independently when the state and intent of each vehicle is shared. Command terminals can globally coordinate the behavior of multiple vehicles and then send instructions to corresponding individual.

\begin{figure}[!t]
	\centering
	\includegraphics[width=3.5in]{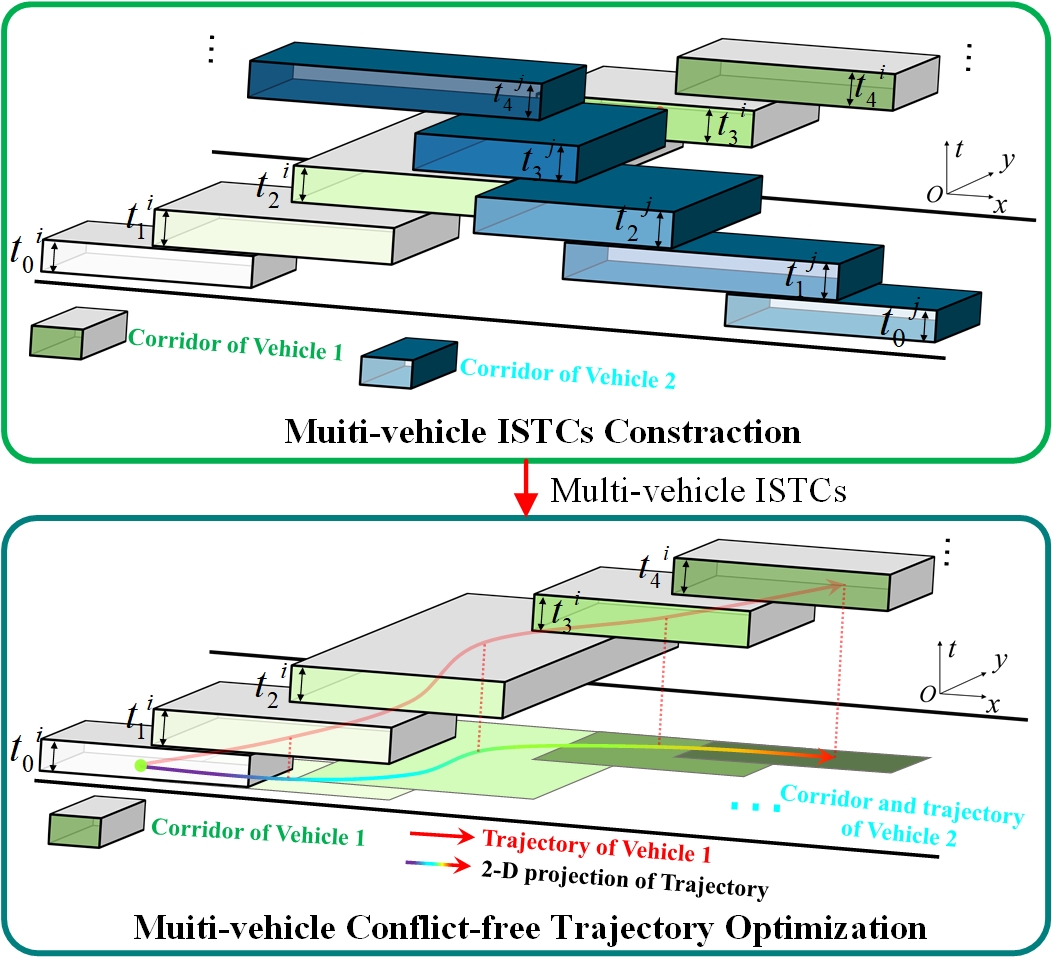}
	\caption{Flowchart of the proposed hierarchical method.}
	\label{fig:Flowchat}
	\vspace{-1.0em}
\end{figure}
The greatest challenge for multi-vehicle coordinated motion planning is how to ensure no-conflicts for all vehicles in space-time configuration space. The current research on multi-vehicle planning mainly focus on specific scenarios \cite{rios2016survey}, such as merging and intersection. A universal method to ensure the safety and efficiency of multi-vehicle driving in common scenarios is still in deficiency. Dedicated to changing this situation, this paper proposes a hierarchical planning method based on interactive spatio-temporal corridors (ISTCs), which provide a unified representation of vehicles and other environmental elements and ensures safety while efficiently avoid conflicts.

As shown in Fig.\ref{fig:Flowchat}, a novel hierarchical framework is proposed to accomplish the multi-vehicle coordinated motion planning in general scenes. In the first layer of the main framework, Mixed Integer Quadratic Programming (MIQP) is designed to construct ISTCs capable of resolving conflicts in generic multi-vehicle scenarios. ISTCs consist of spatial temporal corridors of all vehicles that do not overlap with each other, and the corridor of each vehicle is composed of corridor-cubes on a sequence of consecutive time units. Then, for the second layer, Non-Linear Programming (NLP) is settled to generate in-corridor trajectories that satisfy the vehicle dynamics.

The main contributions of the paper are as follows:
\begin{itemize}
	\item{A novel hierarchical framework for handling multi-vehicle coordinated motion planning is proposed. This framework can decouple the multi-vehicle coordinated motion planning problem into single-vehicle trajectory optimization problems under specified constraints, reducing the overall complexity and improving the efficiency in general scenarios}
	\item{An interactive spatio-temporal coupling corridor is established to achieve conflict-free zoning of each vehicle in the spatio-temporal domain. The strong non-convex conflict resolution problem in multi-vehicle scenarios can be solved effectively by setting appropriate constraints and optimization objectives in MIQP.}
	\item{Based on the generated interactive spatio-temporal corridor, each vehicle independently and optimally generates the desired trajectory in the decoupled spatio-temporal domain. Optimal trajectory generation takes into account corridor boundary constraints and vehicle motion characteristic constraints, and generates high quality trajectory timing points in a conflict decoupled space.}
\end{itemize}

The remainder of this paper is organized as follows. Section II presents the literature review and related works. Section III describes the framework of the proposed algorithm. Section IV describes the initial guided trajectory generation for the unstructured environment, the construction of ISTCs, and the final trajectory generation. Section V demonstrates and discusses the results of the ISTCs as well as the final trajectory simulation. Finally, the conclusion and future work are given in Section VI.

\section{Related Work}
In this section, we review the current research status of coordinated motion planning in the fields of both robots and vehicles and present the shortcomings of current methods. In addition, an overview of the application of spatial temporal corridor in trajectory optimization is also provided.

\subsection{Multi-Agent Path Finding for Mobile Robotics}
How to find collision-free trajectories for multi-robot systems is considered as Multi-Agent Path Finding (MAPF) problem and be widely researched for many years. AS a NP-hard problem \cite{yu2013structure}, various methods are proposed to get the trajectories for each participant to achieve their target effectively. Decoupling methods often achieves the shortest computation time, which typically  choose one agent to plan the path every time, ensuring that the chosen agent avoids conflicts with the plans of other agents \cite{silver2005cooperative, barer2014suboptimal, le2019multi}. However, the huge loss of solution space caused by decoupling is unacceptable in real scenes. To alleviate this problem, method based on rules \cite{khorshid2011polynomial}, A* \cite{wagner2011m, standley2010finding, phillips2011sipp}, Conflict Based Search (CBS) \cite{andreychuk2021improving, wen2022cl, li2020efficient} are used to find paths for multiple agents simultaneously to realize asymptotically optimal solutions. In recent years, learning-based approaches have also been widely used in MAPF. For example, Reijnen et al. used Reinforcement Learning (RL) to obtain heuristic values \cite{reijnen2020combining}, Sinkar et al. adopted distributed Deep Learning (DL) models to deal with dynamic objects \cite{sinkar2021multi} and Bai et al. combined the Long Short-Term Memory (LSTM) and Deep Reinforcement Learning (DRL) to accomplish safety in formation control \cite{bai2021learning}. However, although in simple task scenarios the above methods are efficient in generating multi-agent trajectories, most of them greatly simplify or even have no regard for vehicle models and rarely consider the restraints of traffic scenarios. Hence, more details about vehicle characteristics and environmental constraints should be considered if the method is applied to real vehicles on the road.

\subsection{Multi-Vehicle Coordinated Motion Planning}
Multi-vehicle coordinated motion planning has been extensively studied in various scenarios. In the intersection scene, the method Based on multi-vehicle formation control \cite{cai2022multi}, Semi-Stochastic Potential Fields \cite{wang2020learning}, and Partially Observable Markov Decision Process (POMDP) \cite{xia2022interactive} are used to realize the cooperation of vehicles. In the lane change scenario, Reinforcement Learning (RL) combined with opponent modeling network is used to achieve efficient implementation \cite{liang2022hierarchical}. And Li et al. decentralized the lane change maneuver in two stages to balancing computation speed and quality \cite{li2018balancing}. For overtaking scenarios, a method based on artificial potential field method combined with formation control \cite{xie2022distributed} and a cooperative avoidance scheme based on distance estimation strategy \cite{deng2019cooperative} are applied to keep vehicles safe. In addition, to deal with on-ramp coordination, artificial neural network (ANN) combined with Deep Reinforcement Learning (DRL) were proposed to calculate longitudinal acceleration \cite{9557770}, Control Barrier Functions (CBFs) and Control Lyapunov Functions (CLFs) were integrated to realize decentralized control \cite{9564646} and the SAT (Satisfiability) solver \cite{nakamura2020short} were tested with good results. However, most of the above methods are designed for specific traffic scenes and cannot be applied to generic multi-vehicle interaction environment. To find conflict-free trajectories for multiple vehicles in general scenarios, the method based on interactive primitive tree finds the optimal mode of coordination through solving the Mixed Integer Linear Programming (MILP)  \cite{wang2021multi, kessler2019cooperative}. But a new shortcoming is that the discretization of motion space will shrink the solution space and reduce the optimality of the planning results.

\subsection{Trajectory Optimization based on Spatial Temporal Corridors}
The spatial temporal corridor is essentially a set of passable areas for a vehicle at each point in the entire timeline. In contrast to planning methods for decoupling paths and speeds that search in the 2-D$(x,y)$ plane, planning methods that apply spatial temporal corridors can search for solutions in a larger 3-D$(x,y,t)$ configuration space, leading to better trajectories under most conditions. The concept of a large convex area of accessible space and the calculation method were presented in \cite{deits2015computing}. Based on this, \cite{liu2017planning} proposed a method to formulate trajectory generation as a Quadratic Program (QP) using the concept of a Safe Flight Corridor (SFC). In the area of autonomous driving, Spatio-temporal Semantic Corridor (SSC) was first proposed in \cite{ding2019safe} to help generate the spatio-temporal trajectory for the vehicle in complex urban environments. Zhang et al. proposed a hierarchical framework consisting of rough search, fine optimization and safety strip-based collision avoidance \cite{zhang2020trajectory}, using spatial temporal corridors to ensure the safety and efficiency of ego-vehicle trajectories on spatio-temporal maps. \cite{zhang2021unified} defined the spatial temporal corridor as a series of voxels with spatio-temporal connectivity, implements behavioural planning by selecting the corresponding voxels. The trapezoidal prism-shaped corridors are introduced for optimization to generate the final trajectory in \cite{deolasee2022spatio, li2022motion}. However, the above methods all use B{\'e}zier curves to optimize the final trajectory, which to some extent weakens the constraints of the vehicle model and may lead to infeasibility of the final solution. Currently, the properties of spatio-temporal corridors in protecting vehicle safety and simplifying planning problems have rarely been utilized by researchers to deal with multi-vehicle collaboration.

The introduction of spatial temporal corridors in multi-vehicle coordinated motion planning has many advantages. The unified expression of the spatial temporal corridor enhances the applicability of the framework, which no longer limits the proposed method to specific scenarios. Besides, in-corridor trajectory optimization makes it possible to fully take into account the kinematic characteristics of vehicles, conducing to higher safety, quality and efficiency.

\section{Framework}
\begin{figure}[!h]
	\centering
	\includegraphics[width=3.5in]{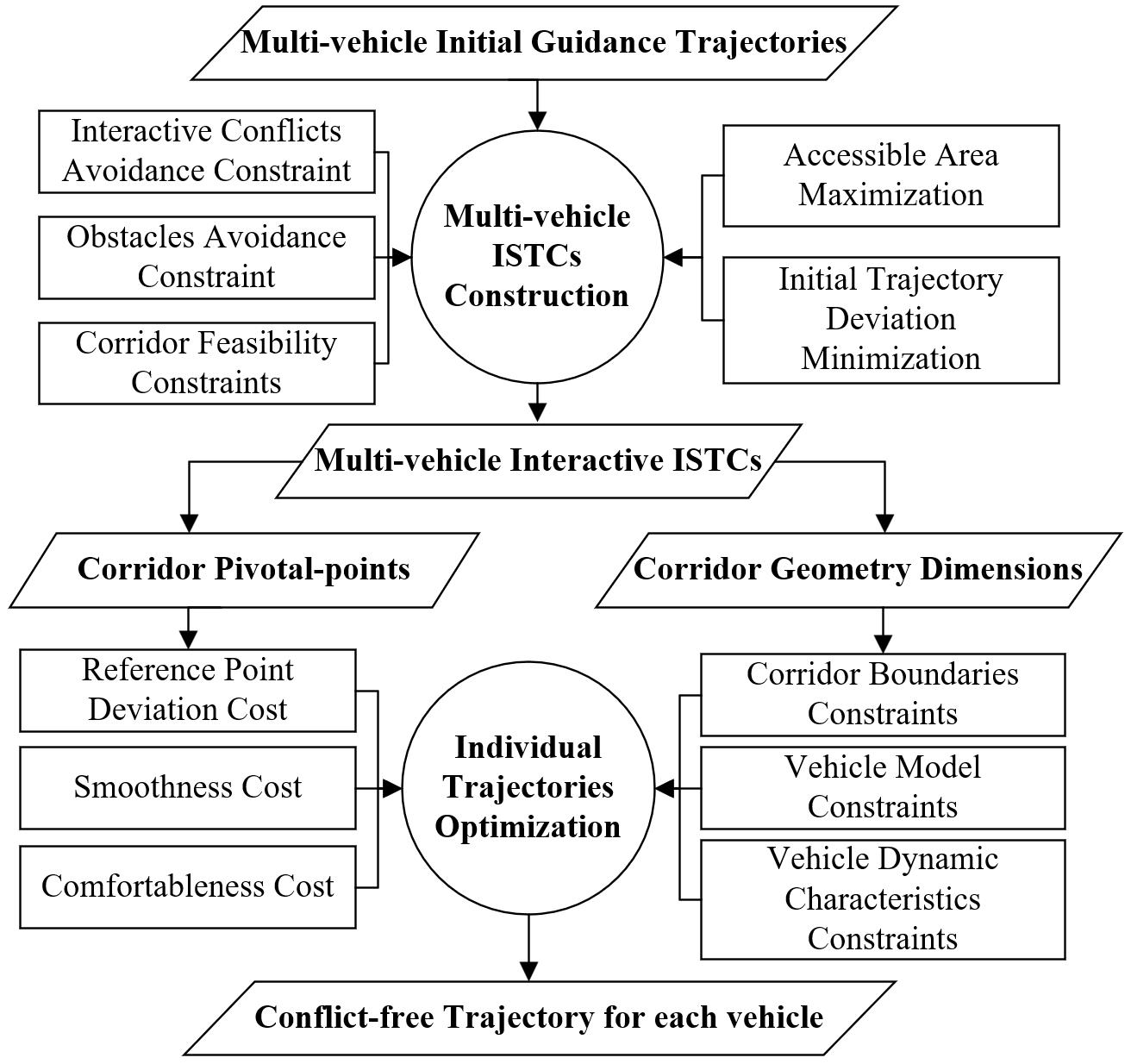}
	\caption{The schematic diagram of the algorithm structure.}
	\label{fig:FrameWork}
\end{figure}

The schematic diagram of the algorithm structure is illustrated in Fig.\ref{fig:FrameWork}. The whole hierarchical approach consists of two main parts: Multi-vehicle ISTCs construction and individual trajectory optimization. In the first part, the central computer globally calculates the ISTCs, i.e. generates a secure channel through the conflict area for each vehicle, after obtaining information about the intentions of all vehicles and environmental constraints. In the second part the vehicles generate dynamically feasible trajectories in their respective spatial temporal corridors, no longer considering interaction conflicts and obstacle collisions.

Prior to the ISTCs calculation, initial guidance trajectories serves as inputs for the intents of vehicles, with the aim of guiding the extension of all spatial temporal corridors. In practice, for structured urban environments, we use lane lines directly as guides, while in other environments, the hybrid A* \cite{dolgov2008practical} algorithm is used to independently find the path of each vehicle satisfying the environmental constraints. After the path is given a constant velocity to complement the time dimension, it serves as the initial guide trajectory of the input. Afterwards, Mixed Integer Linear Programming (MIQP) was settled to constract ISTCs. In the MIQP design, two kinds of conflicts must be considered to ensure safety: 1) Interactive conflicts among vehicles; 2) Environment conflicts between vehicles and obstacles. In addition, a further type of constraints are taken into account to ensure viability: 3) Corridor feasibility constraints. All of the above three constraints can be integrated in the design of MIQP, which are detailed in section IV. As for the objective function, since the accessible area of each vehicle is expected to be maximized and the corridor extension should be kept in line with the driving intentions, we set the optimization objectives as: 1) Maximizing the corridor-cube in each time unit; 2) Minimizing the deviation between pivotal-points of corridor-cubes and initial guidance trajectories.

In the aforementioned module, the constructed ISTC has completely resolved the conflicts and delineated the respective passable areas for each vehicle in the space-time configuration space. Thus, the whole multi-vehicle motion planning problem can be decoupled to single-vehicle trajectory optimization problems. The generation of the final trajectories within the spatial temporal corridors is transformed into a Non-Linear Programming (NLP) problem. Several objectives is considered under multiple constraints for the trajectory optimization. It is noticed that the driving intention information of the initial reference trajectories have been passed to the sequence of pivotal-points in the corridors. Therefore, for ensuring that the trajectories follows vehicles' intents, one of the cost term is to make generated trajectories as close to the pivotal-points as possible. Besides, the quality of trajectory is taken into account, as the cost of comfort and smoothness are indicators to be evaluated. In addition, benefiting from ISTCs, the original strongly non-convex constraint of generating conflict-free trajectories is also transformed into a linear corridor boundary constraint. Finally, vehicle model constraints and vehicle dynamic characteristic constraints are also integrated into the NLP problem with the aim of improving the dynamical applicability of the trajectory.

In this framework, ISTCs construction is to ensure conflict-free and obstacle-free interactions, and trajectory optimization is deliver high-quality executable trajectories. The whole method not only handles the interaction among vehicles well, but also decouples multi-vehicle motion planning into single-vehicle trajectory optimization. On this basis, the final trajectory optimization can be completed on the individual on-board computing platform with much higher real-time performance.

\section{Methodology}
\subsection{Multi-vehicle ISTCs construction}
The initial guidance trajectories haven't taken into account conflicts among vehicles and therefore cannot be performed directly. However, the guidance trajectories provide the basic intent and behavior information of each vehicle, which can be utilized as the reference to demarcate a conflict-free passable area for each vehicle. In this section, how to formulate the mixed integer quadratic programming (MIQP) for constructing ISTCs is described in detail.
\begin{figure}[!h]
	\centering
	\includegraphics[width=3in]{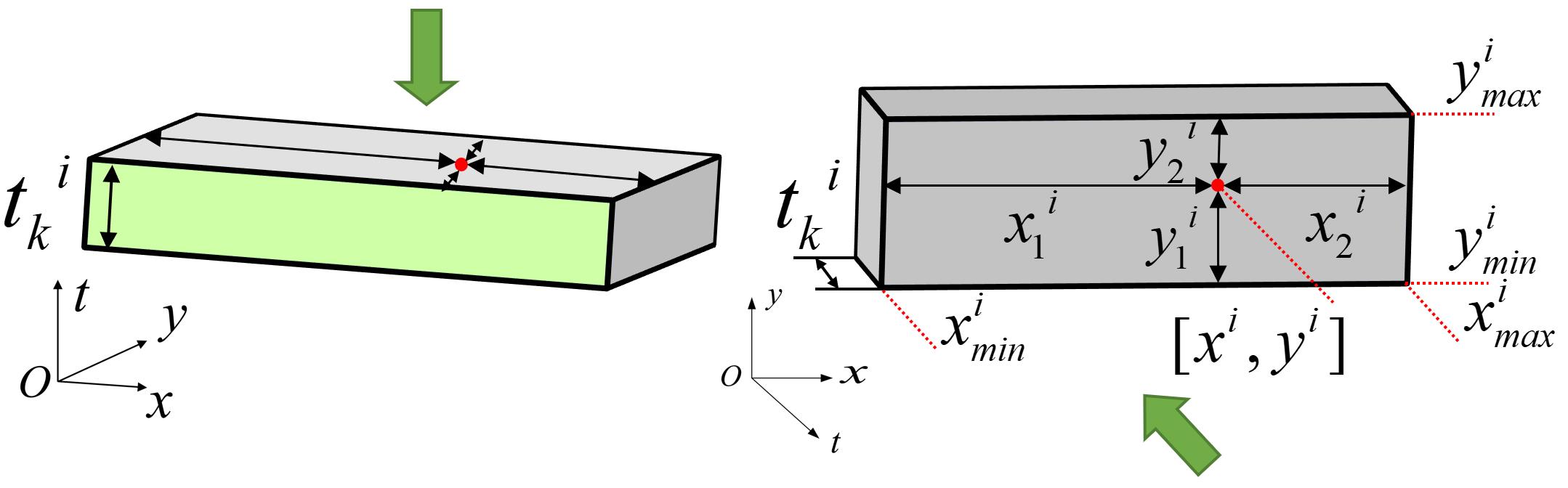}
	\caption{ The structure of a single corridor-cube.}
	\label{fig:Schematic2}
\end{figure}

As the conflicts between vehicles is related to their position at each moment, the spatial temporal corridor needs to be considered in both the time and space domain. The structure of the corridor is showed in Fig.\ref{fig:Flowchat}, which represents the passable area of the vehicles on a sequence of time units. The specific manifestation of one corridor-cube is shown in Fig.\ref{fig:Schematic2}. The position and size of the step for vehicle $i$ in time unit $t_{k}^{i}$ is determined by the pivotal-point ${{p}^{i}}(k)=({{x}_{k}}^{i},{{y}_{k}}^{i})$ and scale vector ${{g}^{i}}(k)=[x_{1}^{i}(k),x_{2}^{i}(k),y_{1}^{i}(k),y_{2}^{i}(k)]$. Hence, each corridor-cube can be represented by:
\begin{equation}
	Cub{{e}^{i}}(k)=[{{x}_{k}}^{i},{{y}_{k}}^{i},x_{1}^{i}(k),x_{2}^{i}(k),y_{1}^{i}(k),y_{2}^{i}(k)]
\end{equation}
Two aspects of performance indicators mainly considered during the construction of ISTCs are listed below:
\begin{equation}
	J_{\text{area }}^{i}=\omega _{\text{area }}^{i}\sum\limits_{k}{\left( x_{1}^{i}(k)+x_{2}^{i}(k)+y_{1}^{i}(k)+y_{2}^{i}(k) \right)}
\end{equation}
\begin{equation}
	J_{\text{ref }}^{i}=\omega _{\text{ref }}^{i}\sum\limits_{k}{\left( {{\left( {{x}^{i}}(k)-x_{\text{ref}}^{i}(k) \right)}^{2}}+{{\left( {{y}^{i}}(k)-y_{\text{ref}}^{i}(k) \right)}^{2}} \right)}
\end{equation}
where $(x_{\text{ref}}^{i}(k),y_{\text{ref}}^{i}(k))$ is the reference point provided by initial reference trajectory, $J_{\text{area }}^{i}$ indicates the breadth of a corridor-cube and $J_{\text{ref }}^{i}$ indicates the deviation between the pivotal-point and initial reference trajectory.

After that, the MIQP model can be formulated as follows:
\begin{equation}
	\begin{aligned}
		& \underset{{{p}^{i}},{{g}^{i}},\forall i\in \mathcal{V}}{\mathop{\operatorname{minimize}}}\,\sum\limits_{i}{{{\eta }^{i}}}\sum\limits_{k}{\left( -J_{\text{area }}^{i}\left( {{g}^{i}}(k) \right)+J_{\text{ref }}^{i}\left( {{p}^{i}}(k) \right) \right)} \\ 
		& \text{     s}\text{.t}\text{.} \\ 
		& \qquad\forall i\in \mathcal{V},j\in \mathcal{V}\backslash \{i\},k\in \mathcal{K},\text{ obs }m\in \mathcal{O}: \\ 
		& \qquad Cub{{e}^{i}}(k)\cap Cub{{e}^{j}}(k)=\emptyset  \\ 
		& \qquad Cub{{e}^{i}}(k)\cap {{O}^{m}}(k)=\emptyset  \\ 
		& \qquad Other \  feasibility \  constrains
	\end{aligned}
\end{equation}
where ${{\eta }^{i}}$ is the priority weight of the vehicle $i$, $\mathcal{V}$ is the set of vehicles, $\mathcal{K}$ is the set of time steps and $\mathcal{O}$ is the set of obstacles, $Cub{{e}^{i}}(k)\cap Cub{{e}^{j}}(k)=\emptyset$ is the constraint to avoid interactive conflicts among vehicles, and $Cub{{e}^{i}}(k)\cap {{O}^{m}}(k)=\emptyset$ is the constraint of environment from obstacles. Besides, in order to make sure the each corridor is dynamic feasible for its vehicle to generate trajectory, serval feasibility constraints  are taken into account. In this paper, the Gurobi solver is used to solve the model and the average time consumption can be maintained at the level of milliseconds.

\subsubsection{Interactive conflicts avoidance constraint}
\begin{figure}[!b]
	\centering
	\subfloat[]{
		\includegraphics[width=3in]{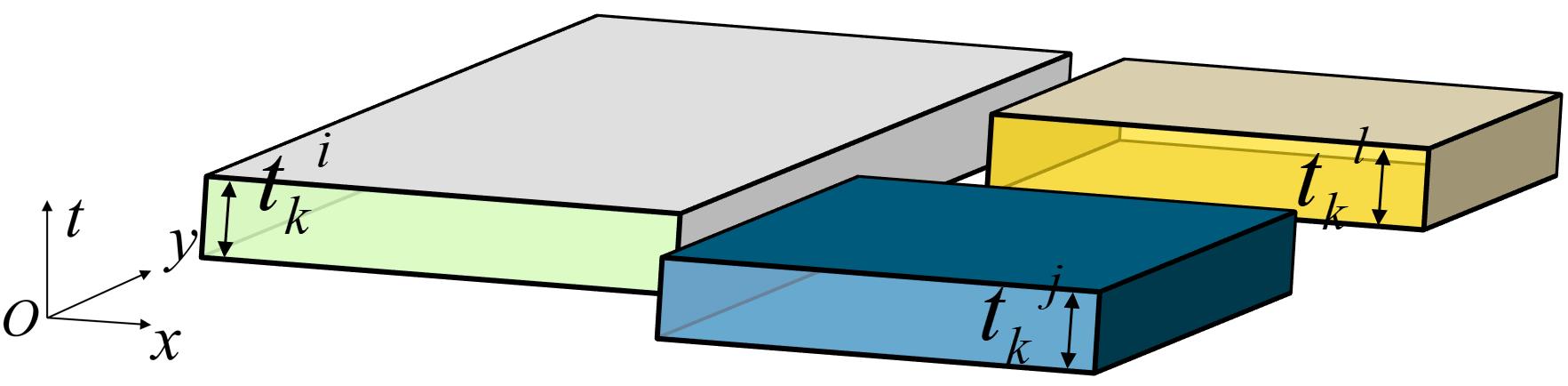}}
	\\
	\subfloat[]{
		\includegraphics[width=3in]{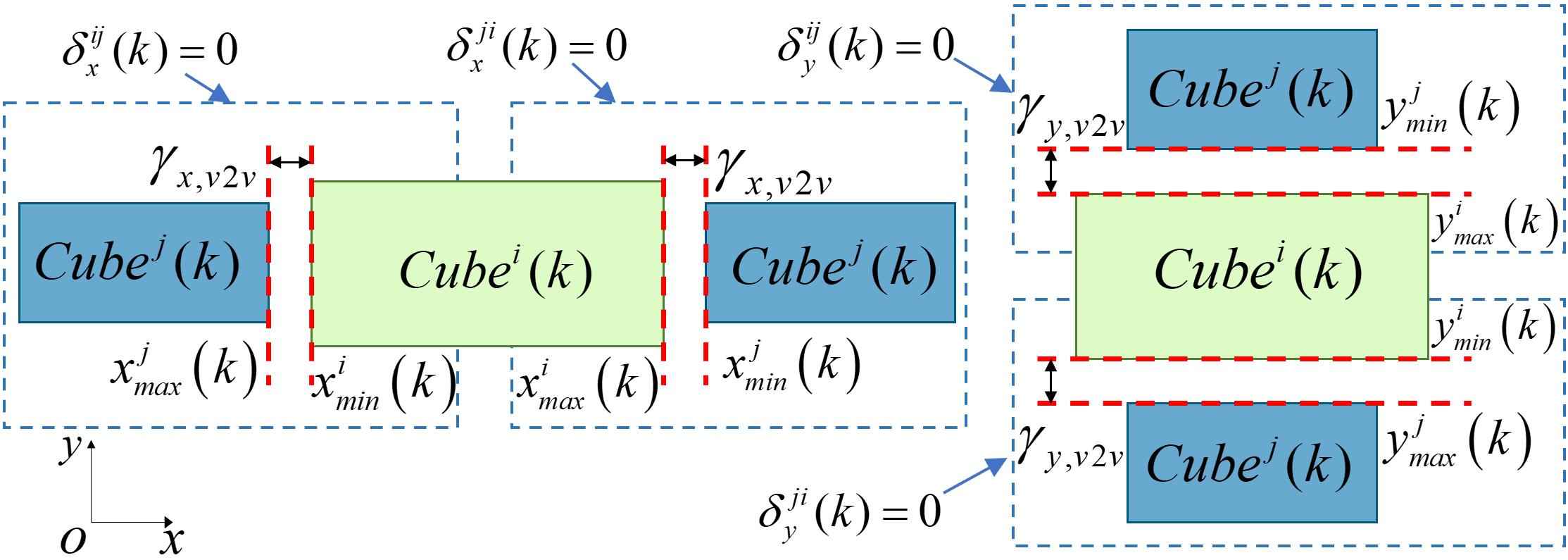}}
	\caption{ (a) An example of three vehicles’ corridor-cubes in a time unit. (b) The projection of corridor-cubes in $x-y$ plane when corresponding constrains is applied.}
	\label{fig:Schematic3}
\end{figure}
This constraint is dedicated to ensure that ISTCs is capable of keeping a safe distance between vehicles, which means that the corridor-cubes of different vehicles in the same time unit can’t overlap each other. In Fig.\ref{fig:Schematic3}(a), the corridor-cubes of three vehicles at the same time unit are shown as an example. In order to ensure the noninterference, the position relationship among corridor-cubes should be considered between every two of them. Therefore, the constraint is formulated as follows:
\begin{equation}
	\left\{ \begin{array}{*{35}{l}}
		x_{min}^{i}(k)-x_{max}^{j}(k)+\delta _{x}^{ij}(k)\cdot M\ge {{\gamma }_{x,v2\text{v}}}  \\
		x_{min}^{j}(k)-x_{max}^{i}(k)+\delta _{x}^{ji}(k)\cdot M\ge {{\gamma }_{x,\text{v2v}}}  \\
		y_{min}^{i}(k)-y_{max}^{j}(k)+\delta _{y}^{ij}(k)\cdot M\ge {{\gamma }_{y,v2\text{v}}}  \\
		y_{min}^{j}(k)-y_{max}^{i}(k)+\delta _{y}^{ji}(k)\cdot M\ge {{\gamma }_{y,v2\text{v}}}  \\
		\delta _{x}^{ij}(k)+\delta _{x}^{ji}(k)+\delta _{y}^{ij}(k)+\delta _{y}^{ji}(k)\le 3  \\
		\delta _{x}^{ij}(k)\in \{0,1\},\delta _{x}^{ji}(k)\in \{0,1\},\\
		\delta _{y}^{ij}(k)\in \{0,1\},\delta _{y}^{ji}(k)\in \{0,1\}  \\
	\end{array} \right.
\end{equation}
where $M$ is the maximal number, $\delta _{x}^{ij}(k), \delta _{x}^{ji}(k), \delta _{y}^{ij}(k), \delta _{y}^{ji}(k)$ are integer variables with the value of 1 or 0, ${{\gamma }_{x,v2\text{v}}}$ and ${{\gamma }_{y,v2\text{v}}}$ are the safety threshold in the $x$ and $y$ directions between vehicles, $[x_{min}^{i}(k),x_{max}^{i}(k),y_{min}^{i}(k),y_{max}^{i}(k)]$ and $[x_{min}^{j}(k),x_{max}^{j}(k),y_{min}^{j}(k),y_{max}^{j}(k)]$ are the boundaries of $Cube^i$ and $Cube^j$ in the same time unit. If one of the integer variables equal to 1, means that the inequality containing this integer variable is always true so that the corresponding constraint is disabled. But if the integer variables are equal to 0, the effect is shown in Fig.\ref{fig:Schematic3}(b). For example, when $\delta _{x}^{ij}(k)$ is equal to 0, the first inequality is true only if the left boundary value $x_{min}^{i}(k)$ of $Cube^i$ is bigger than the right boundary value $x_{min}^{i}(k)$ of $Cube^j$. In the same way, the cases of other three integer variables equal to 0 are also shown in Fig.\ref{fig:Schematic3}(b). At least one of the four integer variables equaling 0 ensures that in each time unit the passable areas of two vehicles are independent of each other. By adding this constraint to all vehicles in pairs, the safety of the entire interaction is ensured.

\subsubsection{Obstacles avoidance constraint}
\begin{figure}[!h]
	\centering
	\includegraphics[width=3in]{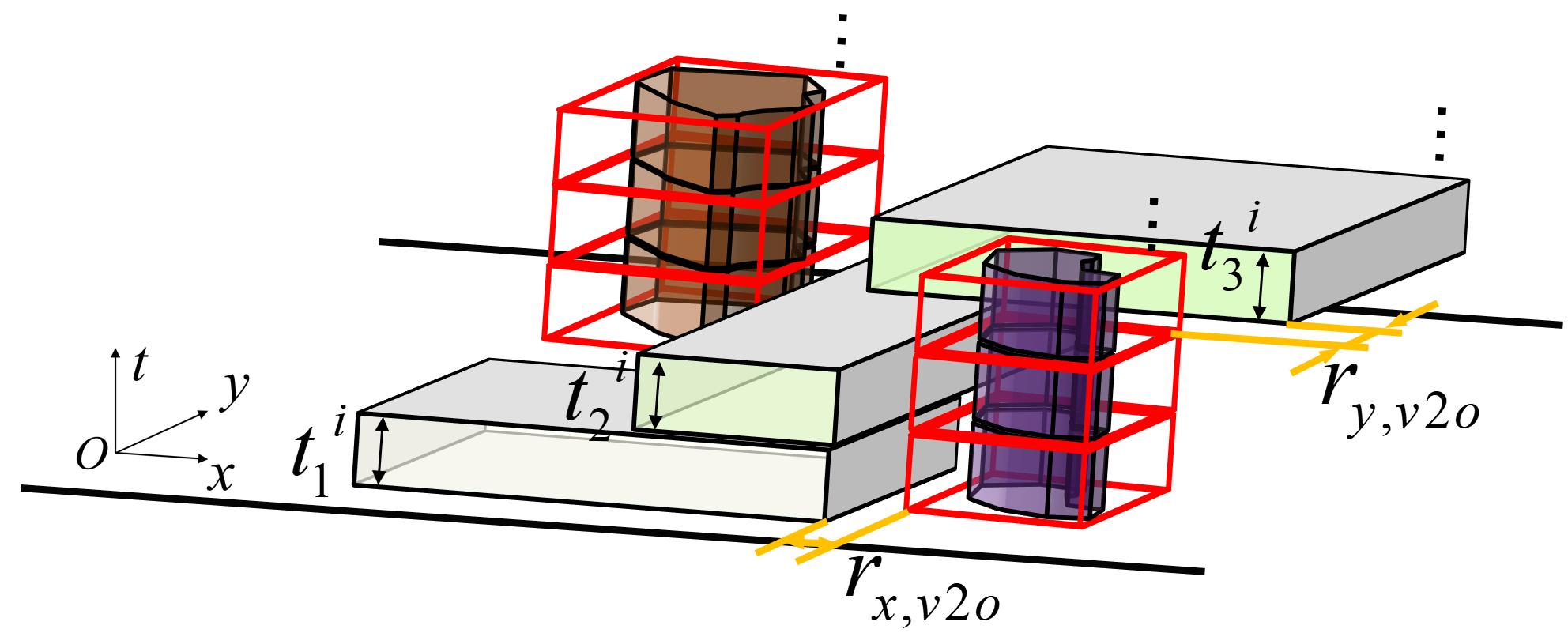}
	\caption{ An example of representing obstacles as rectangles.}
	\label{fig:Schematic4}
\end{figure}

Due to spatial temporal corridor can provide a unified representation of various semantic elements in environments, we can also represent the obstacle as a cube box and use the method similar to (1) to ensure non-conflicts between vehicles and obstacles, as shown in Fig.\ref{fig:Schematic4}. Hence, the constrains can be described as:
\begin{equation}
	\left\{ \begin{array}{*{35}{l}}
		x_{min }^{i}(k)-x{_{max }^{\text{obs }m}}(k)+\delta _{x}^{i,\text{obs }m}(k) \cdot M\ge {{r}_{x,v2o}}  \\
		x{_{min }^{\text{obs }m}}(k)-x{_{max }^{i}}(k)+\delta {_{x}^{\text{obs }m, i}}(k) \cdot M\ge {{r}_{x,v2o}}  \\
		y_{min }^{i}(k)-y{_{max }^{\text{obs }m}}(k)+\delta _{y}^{i,\text{obs }m}(k)M\ge {{r}_{y,v2o}}  \\
		y{_{min }^{\text{obs }m}}(k)-y{_{max }^{i}}(k)+\delta {_{y}^{\text{obs }m,i}}(k) \cdot M\ge {{r}_{y,v2o}}  \\
		\delta _{x}^{i,\text{obs }m}(k)+\delta _{x}^{\text{obs }m,i}(k)+\delta _{y}^{i,\text{obs }m}(k)+\delta _{y}^{\text{obs }m,i}(k)\le 3  \\
		\delta _{x}^{i,\text{obs }m}(k)\in \{0,1\},\delta _{x}^{\text{obs }m,i}(k)\in \{0,1\},\\
		\delta _{y}^{i,\text{obs }m}(k)\in \{0,1\}, \delta _{y}^{\text{obs }m,i}(k)\in \{0,1\}  \\
	\end{array} \right.
\end{equation}
where $[x{{_{\min}^{obs}}^{m}}(k),x{{_{\max}^{obs}}^{m}}(k),y{{_{\max }^{obs}}^{m}}(k),y{{_{\max }^{obs}}^{m}}(k) ]$ is the cube box of obstacle $m$ in time unit $t{_k^i}$, ${{r}_{x,v2o}}$ and ${{r}_{y,v2o}}$ are the safety threshold, $\delta _{x}^{i,\text{obs }m}(k)$, $\delta _{x}^{\text{obs }m,i}(k)$, $\delta _{y}^{i,\text{obs }m}(k)$, $\delta _{y}^{\text{obs }m,i}(k)$ are integer variables.
\subsubsection{Corridor feasibility constraints}
After ensuring the spatial temporal corridor of each vehicle is conflict free with other vehicles and obstacles, we also need to add constraints to ensure that the corridor is executable for the vehicle to generate passable trajectories in subsequent modules. Therefore, we need to take account of vehicle characteristics when construct ISTCs and make sure the vehicle can reach every corridor-cube at the specified time. This kind of constraints are mainly considered in two aspects. One is the dimension and boundary constraints of every single corridor-cube in each time unit. Each corridor-cube should be able to accommodate the car-box, and due to the limitations of vehicle dynamic characteristics, the length of each corridor-cube should not exceed the maximum distance that the vehicle can drive within a unit time period. Another is the dimension and boundary constraints of adjacent corridor-cubes between continuous time periods. Considering the behavioral continuity of vehicles, the adjacent corridor-cubes need to have a minimum overlap to make sure the vehicle can transition from the current corridor-cube to the next. Meanwhile, the distance between pivotal-points of adjacent corridor-cubes should be limited according to the maximum speed of vehicles.
\begin{figure}[!h]
	\centering
	\includegraphics[width=3in]{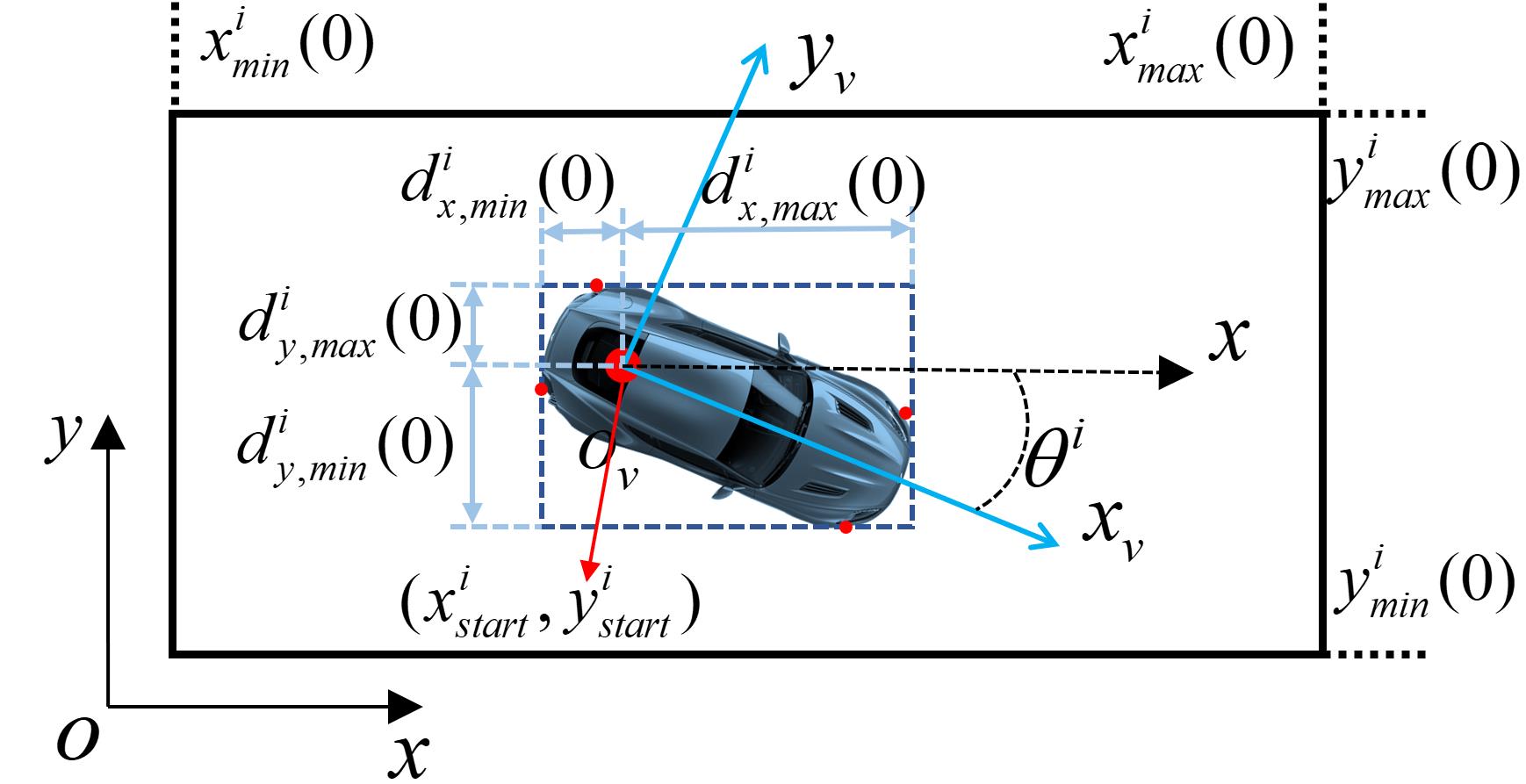}
	\caption{ $x-y$ plane projections of the car-box and corridor-cube at the initial moment}
	\label{fig:Schematic5}
\end{figure}
The dimension and boundary constraints of the corridor-cube in a time unit is firstly considered. Each corridor-cube should have enough space to encase the car-box of vehicle with uncertain orientation.
\begin{equation}
	\left\{ \begin{array}{*{35}{l}}
		x_{1}^{i}(k)+x_{2}^{i}(k)\ge \gamma _{x,car}^{i}  \\
		y_{1}^{i}(k)+y_{2}^{i}(k)\ge \gamma _{y,car}^{i}  \\
	\end{array} \right.
\end{equation}
where $\gamma _{x,car}^{i},\gamma _{y,car}^{i}$ are the minimum distance thresholds along $x$ and $y$ axes required to accommodate the vehicle geometry in the every time period. In practical, we set $\gamma _{x,car}^{i}=\gamma _{y,car}^{i}=\sqrt{{{({{l}^{i}})}^{2}}+{{({{w}^{i}})}^{2}}}$, where ${{l}^{i}}$ and ${w}^{i}$ are the length and width of vehicle. Besides, since the initial state of vehicles is deterministic and non-adjustable, the corresponding corridor-cube has more stringent constraints, as shown in Fig.\ref{fig:Schematic5}. The following constraint is applied:
\begin{equation}
	\left\{ \begin{aligned}
		& x_{min}^{i}(0)\le x_{start}^{i}+{{d}_{x,min}}(0)\\
		& x_{max}^{i}(0)\ge x_{start}^{i}+{{d}_{x,max}}(0) \\ 
		& y_{min}^{i}(0)\le y_{start}^{i}+{{d}_{y,min}}(0)\\
		& y_{max}^{i}(0)\ge y_{start}^{i}+{{d}_{y,max}}(0) \\ 
	\end{aligned} \right.
\end{equation}
where $(x_{start}^{i},y_{start}^{i})$ is the initial position of the vehicle, $[x_{min}^{i}(0), x_{max}^{i}(0),  y_{min}^{i}(0), y_{max}^{i}(0)]$ are the boundary of the start corridor-cube in $x$ and $y$ directions, ${{d}_{x,min}}(0)$, ${{d}_{x,max}}(0)$, ${{d}_{y,min}}(0)$, ${{d}_{y,max}}(0)$ are the distance between the edge of the car-box and the position of the vehicle that can be calculated  from the state and geometry of the vehicle. In addition, due to the limitations of the maximum speed and acceleration of the vehicle, the dimensions of each corridor-cube in a unit time period are limited. The driving range constraint in time unit $t_{k}^{i}$ of vehicle $i$ is as follows:
\begin{equation}
	\left\{ \begin{aligned}
		& x_{min}^{i}(k)\ge \alpha _{x}^{i}\cdot x_{min,drive}^{i}(k)\\
		& {{x}_{max}}(k)\le \alpha _{x}^{i}\cdot x_{max,drive}^{i}(k) \\ 
		& y_{min}^{i}(k)\ge \alpha _{y}^{i}\cdot y_{min,drive}^{i}(k)\\
		& y_{max}^{i}(k)\le \alpha _{y}^{i}\cdot y_{max,drive}^{i}(k) \\ 
	\end{aligned} \right.
\end{equation}
where $x_{min,drive}^{i}(k)$, $ x_{max,drive}^{i}(k)$, $y_{min,drive}^{i}(k)$, $y_{max,drive}^{i}(k)$ are the minimum and maximum boundaries of vehicle driving range in the $x$ and $y$ directions. Besides, notice that the relaxation factor $\alpha _{x}^{i},\alpha _{y}^{i}$ is set to increase the success rate during solving the MIQP problem. Usually we set $\alpha _{x}^{i}$ and $\alpha _{y}^{i}$ equal to 2. The four driving range values can be calculated by formulation bellow:
\vspace{-1em}
\begin{equation}
	\begin{aligned}
		& x_{min,drive}^{i}(k)=\\
		&\quad \left\{ \begin{array}{*{35}{l}}
			{{x}^{i}}(k-1)+\gamma _{s-}^{i}\cos \theta _{\text{ref }}^{i}(k-1),\text{ if }\cos {{\theta }^{i}}(k-1)\ge 0  \\
			{{x}^{i}}(k-1)+\gamma _{s+}^{i}\cos \theta _{\text{ref }}^{i}(k-1),\text{ else }  \\
		\end{array} \right. \\ 
		& x_{max,drive}^{i}(k)=\\
		&\quad \left\{ \begin{array}{*{35}{l}}
			{{x}^{i}}(k-1)+\gamma _{s-}^{i}\cos \theta _{\text{ref }}^{i}(k-1),\text{ if }\cos {{\theta }^{i}}(k-1)\le 0  \\
			{{x}^{i}}(k-1)+\gamma _{s+}^{i}\cos \theta _{\text{ref }}^{i}(k-1),\text{ else }  \\
		\end{array} \right. \\ 
		& y_{min,drive}^{i}(k)=\\
		&\quad \left\{ \begin{array}{*{35}{l}}
			{{y}^{i}}(k-1)+\gamma _{s-}^{i}\sin \theta _{\text{ref }}^{i}(k-1),\text{ if }\sin {{\theta }^{i}}(k-1)\ge 0  \\
			{{y}^{i}}(k-1)+\gamma _{s+}^{i}\sin \theta _{\text{ref }}^{i}(k-1),\text{ else }  \\
		\end{array} \right. \\ 
		& y_{max,drive}^{i}(k)=\\
		&\quad \left\{ \begin{array}{*{35}{l}}
			{{y}^{i}}(k-1)+\gamma _{s-}^{i}\sin \theta _{\text{ref }}^{i}(k-1),\text{ if }\sin {{\theta }^{i}}(k-1)\le 0  \\
			{{y}^{i}}(k-1)+\gamma _{s+}^{i}\sin \theta _{\text{ref }}^{i}(k-1),\text{ else }  \\
		\end{array} \right. \\  
	\end{aligned}
\end{equation}

where $\gamma _{s+}^{i}$, $\gamma _{s-}^{i}$ are the maximum distance threshold of vehicle $i$ driving forward and backward along the current heading angle, ${{\theta }^{i}}(k-1)$ is the heading angle in time unit $t_{k-1}^{i}$ obtained according to the initial guidance trajectory
\begin{figure}[!h]
	\centering
	\subfloat[]{
		\includegraphics[width=2in]{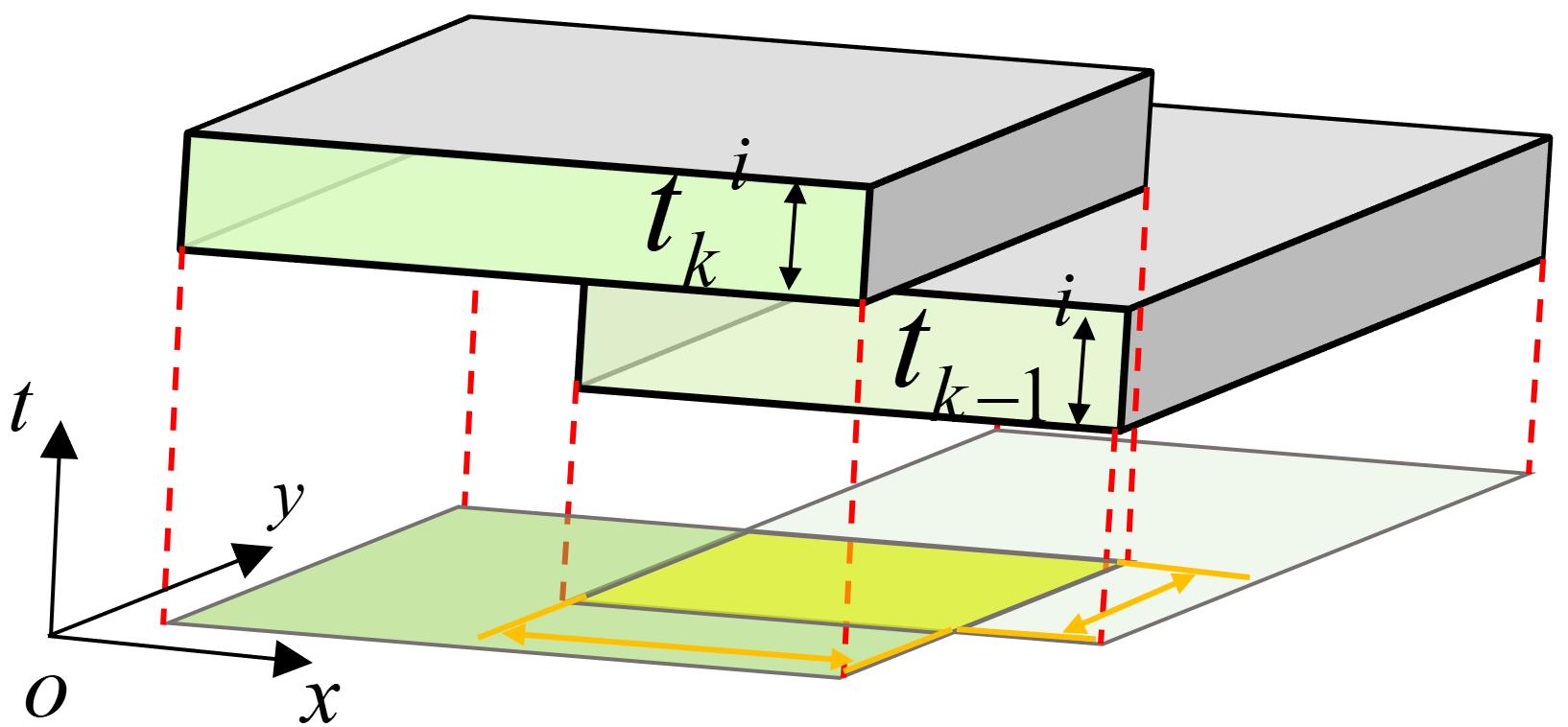}}
	\\
	\subfloat[]{
		\includegraphics[width=1.6in]{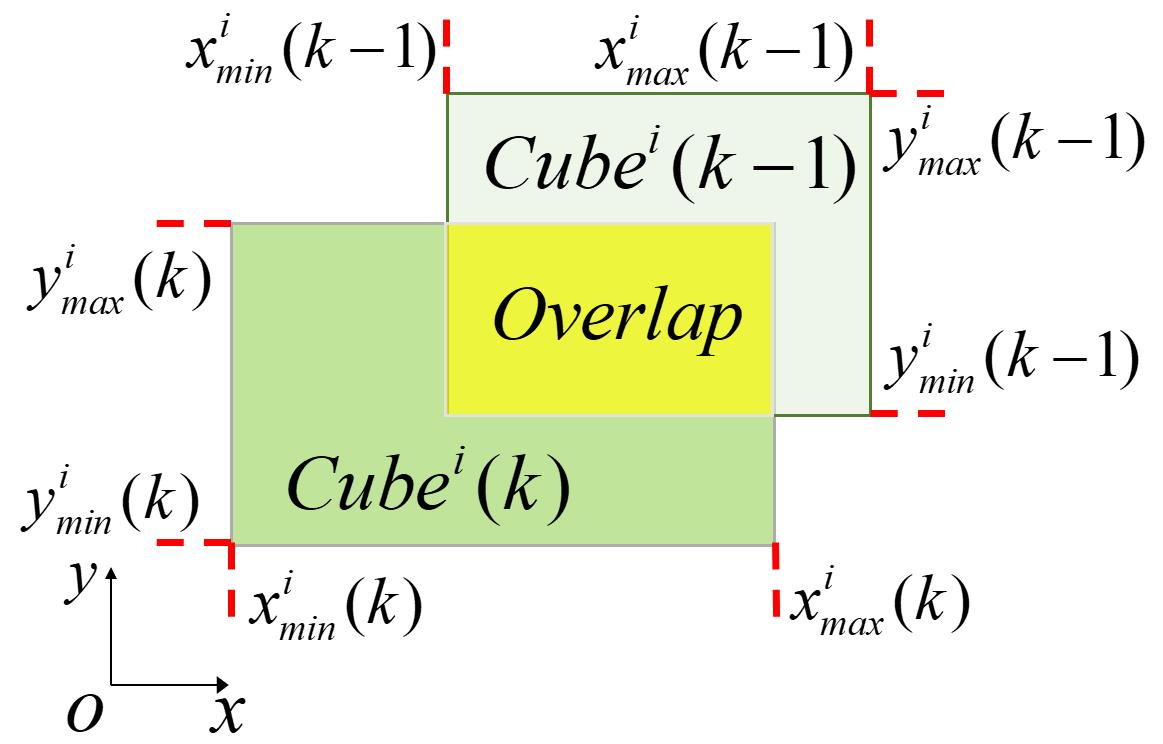}}
	\quad
	\subfloat[]{
		\includegraphics[width=1.2in]{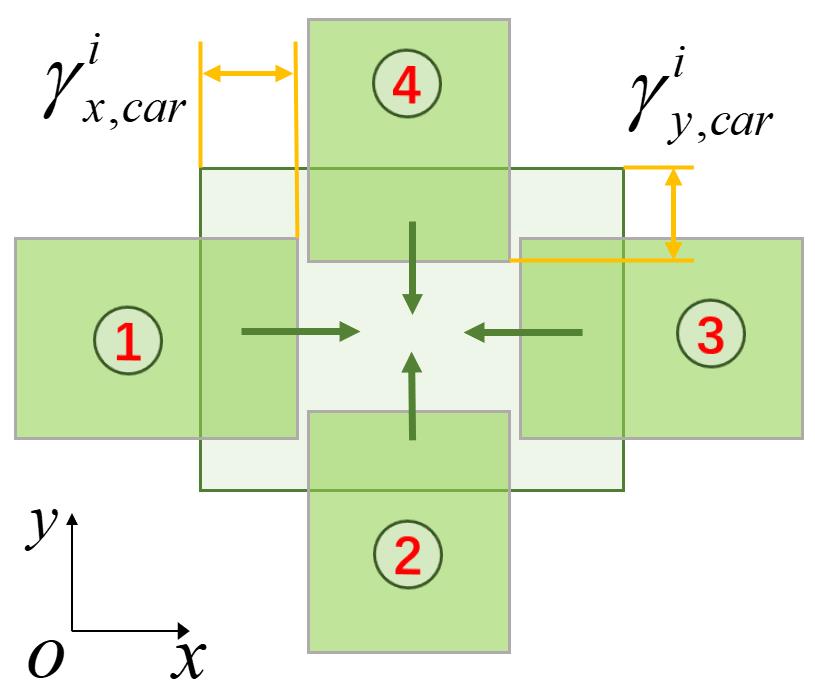}}
	\caption{ (a) The overlap constraints between adjacent corridor-cubes; (b) The 2-D $x-y$ plane projection of overlap between two adjacent corridor-cubes; (c) Four basic cases of overlap between adjacent corridor-cubes.}
	\label{fig:Schematic6}
\end{figure}
The dimension and boundary constraints of adjacent corridor-cubes between continuous time periods also need to be considered. In order to ensure that vehicles transition between successive corridor-cubes are feasible, a minimum overlap requirement for two consecutive corridor-cubes should be imposed, as reflected in Fig.\ref{fig:Schematic4}(a). In the $x$-$y$ plane projection shown in Fig.\ref{fig:Schematic4}(b), we can more intuitively observe the overlapping relationship between the two corridor-cubes. Sufficient overlapping area between two corridors-cubes will be guaranteed as long as position relation is constrained along $x$ and $y$ directions respectively. In Fig.\ref{fig:Schematic4}(c), four basic situations of overlaps are shown with the arrows pointing in the direction of increasing imbricate areas. Any other overlapping cases that meets the requirements is evolved from the four basic cases. The constraints can be described as follows:
\begin{equation}
	\left\{ \begin{aligned}
		& x_{max}^{i}(k)-x_{min}^{i}(k-1)\ge \gamma _{x,car}^{i} \\ 
		& x_{min}^{i}(k)-x_{max}^{i}(k-1)\le \gamma _{x,car}^{i} \\
		& {{y}_{max}}(k)-{{y}_{min}}(k-1)\ge \gamma _{y,car}^{i} \\ 
		& y_{min}^{i}(k)-y_{max}^{i}(k-1)\le \gamma _{y,car}^{i} \\ 
	\end{aligned} \right.
\end{equation}
where $\gamma _{x,car}^{i},\gamma _{y,car}^{i}$ have the same meaning as the equation (7). Finally, the traveling distance between the pivotal-points of the continuous corridor-cubes is limited by the maximum speed of vehicles, which can be described as follows:
\begin{equation}
	\left\{ \begin{aligned}
		& |{{x}^{i}}(k)-{{x}^{i}}(k-1)|\le |\gamma _{s+}^{i}\cos {{\theta }_{ref}}(k-1)| \\ 
		& |{{y}^{i}}(k)-{{y}^{i}}(k-1)|\le |{{\gamma }_{s+}}\sin \theta _{ref}^{i}(k-1)| \\ 
	\end{aligned} \right.
\end{equation}
where $|{{x}^{i}}(k)-{{x}^{i}}(k-1)|$ and $|{{y}^{i}}(k)-{{y}^{i}}(k-1)|$ are the distance between two adjacent pivotal-points in the $x$ and $y$ directions and this formula restricts them smaller than the component of the maximum forward distance on  both $x$ and $y$ axis.

\subsection{Multi-Vehicle Conflict-free Trajectory generation}
After the construction of ISTCs, the whole multi-vehicle planning problem is decoupled into single vehicle trajectory planning problems. Two features of each vehicle's spatial temporal corridor play a large role in the trajectory optimization, namely the pivotal-points and corridor boundaries. The former is utilized to guide the the direction of optimal trajectory, while the latter can constrain the trajectory within a safe and feasible driving area. Since each corridor is constructed in the three-dimensional space-time configuration space, we can use the lateral and longitudinal coupling method to solve the trajectory optimization problem and generate the trajectory in the spatio-temporal map directly.

In this module three cost will be considered to get the final trajectory: cost of smoothness $J_{smooth}^{i}$, cost of comfort $J_{comfort}^{i}$ and cost of deviation from pivotal-point $J_{pivotal}^{i}$. The cost of smoothness is related with the curvature $\kappa _{t}^{i}$ of trajectory:
\begin{equation}
	J_{{smooth }}^{i}=\omega _{\kappa }^{i}\sum\limits_{t}{\kappa _{t}^{i2}}
\end{equation}
where $\omega _{\kappa }^{i}$ is the smoothness weight of vehicle $i$. The cost of comfort is consisted of longitudinal and lateral parts:
\begin{equation}
	J_{{comfort }}^{i}=\omega _{\beta }^{i}\sum\limits_{t}{{{\left( \beta _{t}^{i} \right)}^{2}}}+\omega _{j}^{i}\sum\limits_{t}{{{\left( j_{t}^{i} \right)}^{2}}}
\end{equation}
where $\beta _{t}^{i}$ is the change rate of front wheel angle with the weight of $\omega _{\beta }^{i}$, and $j_{t}^{i}$ is the change rate of the acceleration with the weight of $\omega _{j}^{i}$. The cost of deviation from pivotal-points is as follows:
\begin{equation}
	\begin{aligned}
		J_{pivotal}^{i}& = \omega _{{px }}^{i}\sum\limits_{k,t\in \left[ {{t}_{k}},t_{k+1}^{i} \right)}{{{\left( x_{t}^{i}-{{x}^{i}}(k) \right)}^{2}}} \\
		&\qquad + \omega _{{py }}^{i}\sum\limits_{k,t\in \left[ \in {{t}_{k}},t_{k+1}^{i} \right)}{{{\left( y_{t}^{i}-{{y}^{i}}(k) \right)}^{2}}}
	\end{aligned}
\end{equation}
where $\omega _{{px }}^{i}$ and $\omega _{{py }}^{i}$ represent the deviation weights of the vehicle $i$ from the pivotal-point in the $x$ and $y$ directions, respectively.

Then, the NLP problem can be formulated:
\begin{equation}
	\begin{aligned}
		& \underset{\beta _{t}^{i},j_{t}^{i}}{\mathop{\operatorname{minimize}}}\,\left( J_{{smooth }}^{i}+J_{{comfort }}^{i}+J_{pivotal}^{i} \right) \\ 
		& \quad s.t. \\ 
		& \qquad \forall vehicle\text{ }i,time\text{ }t,\beta _{t}^{i}\in \mathbb{R},j_{t}^{i}\in \mathbb{R} \\ 
		& \qquad S_{t+1}^{i}=f(S_{t}^{i},\beta _{t}^{i},j_{t}^{i}) \\ 
		& \qquad \sum\limits_{t=0}^{t}{\beta _{t}^{i}}\in [-\delta _{\max }^{i},\delta _{\max }^{i}],\sum\limits_{t=0}^{t}{j_{t}^{i}}\in [-a_{\text{dec}\max }^{i},a_{\text{acc}\max }^{i}] \\ 
		& \qquad (x_{t}^{i},y_{t}^{i})\in {{Corridor}^{i}} \\ 
	\end{aligned}
\end{equation}
where $[-\delta _{\max }^{i},\delta _{\max }^{i}]$ is the range of front wheel steering Angle, $[-a_{\text{dec}\max }^{i},a_{\text{acc}\max }^{i}]$ is the range of acceleration, $S_{t}^{i}$ is the state parameters set of vehicle $i$ at time $t$, $S_{t+1}^{i}=f(S_{t}^{i},\beta _{t}^{i},j_{t}^{i})$ is the state transition at each timestep with the equality constraints of the vehicle model, and can be formulated as follows:
\begin{equation}
	\left[ \begin{matrix}
		x_{t+1}^{i}  \\
		y_{t+1}^{i}  \\
		\theta _{t+1}^{i}  \\
		\delta _{t+1}^{i}  \\
		v_{t+1}^{i}  \\
		a_{t+1}^{i}  \\
	\end{matrix} \right]=\left[ \begin{matrix}
		x_{t}^{i}+v_{t}^{i}\cos \theta _{t}^{i}\Delta t  \\
		y_{t}^{i}+v_{t}^{i}\sin \theta _{t}^{i}\Delta t  \\
		\theta _{t}^{i}+v_{t}^{i}\frac{\tan \delta _{t}^{i}}{L}\Delta t  \\
		\delta _{t}^{i}  \\
		v_{t}^{i}+a_{t}^{i}\Delta t  \\
		a_{t}^{i}  \\
	\end{matrix} \right]+\left[ \begin{matrix}
		0 & 0  \\
		0 & 0  \\
		0 & 0  \\
		\Delta t & 0  \\
		0 & 0  \\
		0 & \Delta t  \\
	\end{matrix} \right]\left[ \begin{matrix}
		\beta _{t}^{i}  \\
		j_{t}^{i}  \\
	\end{matrix} \right]
\end{equation}
the equation above is a discretized kinematic bicycle model of wheeled vehicles, which is assumed that the vehicle is driving on a relatively flat road without considering the vertical and roll motion. The vehicle dynamic characteristics constraints of the vehicle, such as the maximum acceleration and the maximum front wheel steering angle are described as:
\begin{equation}
	\left\{ \begin{array}{*{35}{l}}
		-\delta _{max }^{i}\le \delta _{t}^{i}\le \delta _{max }^{i}  \\
		-a_{\text{dec }max }^{i}\le a_{t}^{i}\le a_{\text{acc }max }^{i}  \\
	\end{array} \right.
\end{equation}
Besides, $(x_{t}^{i},y_{t}^{i})\in {Corridor}^{i}$ is the constraint to keep the trajectory inside the spatial temporal corridor, which means that the car-box of the vehicle should be enclosed by corridor-cubes of its corridor at all times, similar to the start position shown in Fig.\ref{fig:Schematic5}, the constraint can be described as:
\begin{equation}
	\left\{
	\begin{array}{*{35}{l}}
		x_{min }^{i}(k)\le x_{t,{rr}}^{i}\le x_{max }^{i}(k),y_{min }^{i}(k)\le y_{t,{rr}}^{i}\le y_{max }^{i}(k)  \\
		x_{min }^{i}(k)\le x_{t,{fl}}^{i}\le x_{max }^{i}(k),y_{min }^{i}(k)\le y_{t,{fl}}^{i}\le y_{max }^{i}(k)  \\
		x_{min }^{i}(k)\le x_{t,{rl}}^{i}\le x_{max }^{i}(k),y_{min }^{i}(k)\le y_{t,{rl}}^{i}\le y_{max }^{i}(k)  \\
		x_{min }^{i}(k)\le x_{t,{rr}}^{i}\le x_{max }^{i}(k),y_{min }^{i}(k)\le y_{t,{rr}}^{i}\le y_{max }^{i}(k)  \\
	\end{array} \right.
\end{equation}

\section{Experimental Results and Discussion}
In this section, we organized several groups of simulation experiments in unsignalized intersection scenario as well as in challenging dense scenarios. In all experiments, we discretized time in seconds to solve MIQP problem for constructing ISTCs, i.e. each corridor-cube represents a passable area in one second.

The proposed algorithm was implemented in C++ and executed on a laptop running Ubuntu 18.04 with Intel i7-8700H @2.30 GHz CPU and 16 GB of RAM. We used lpsolve solver and Gurobi solver to solve the MIQP of ISTCs constructions and use IPOPT to solve the NLP for generating final trajectories inside each spatial temporal corridor. Besides, GridMap was used to represent the occupancy map of the environment.
\begin{figure}[!h]
	\centering
	\includegraphics[width=2in]{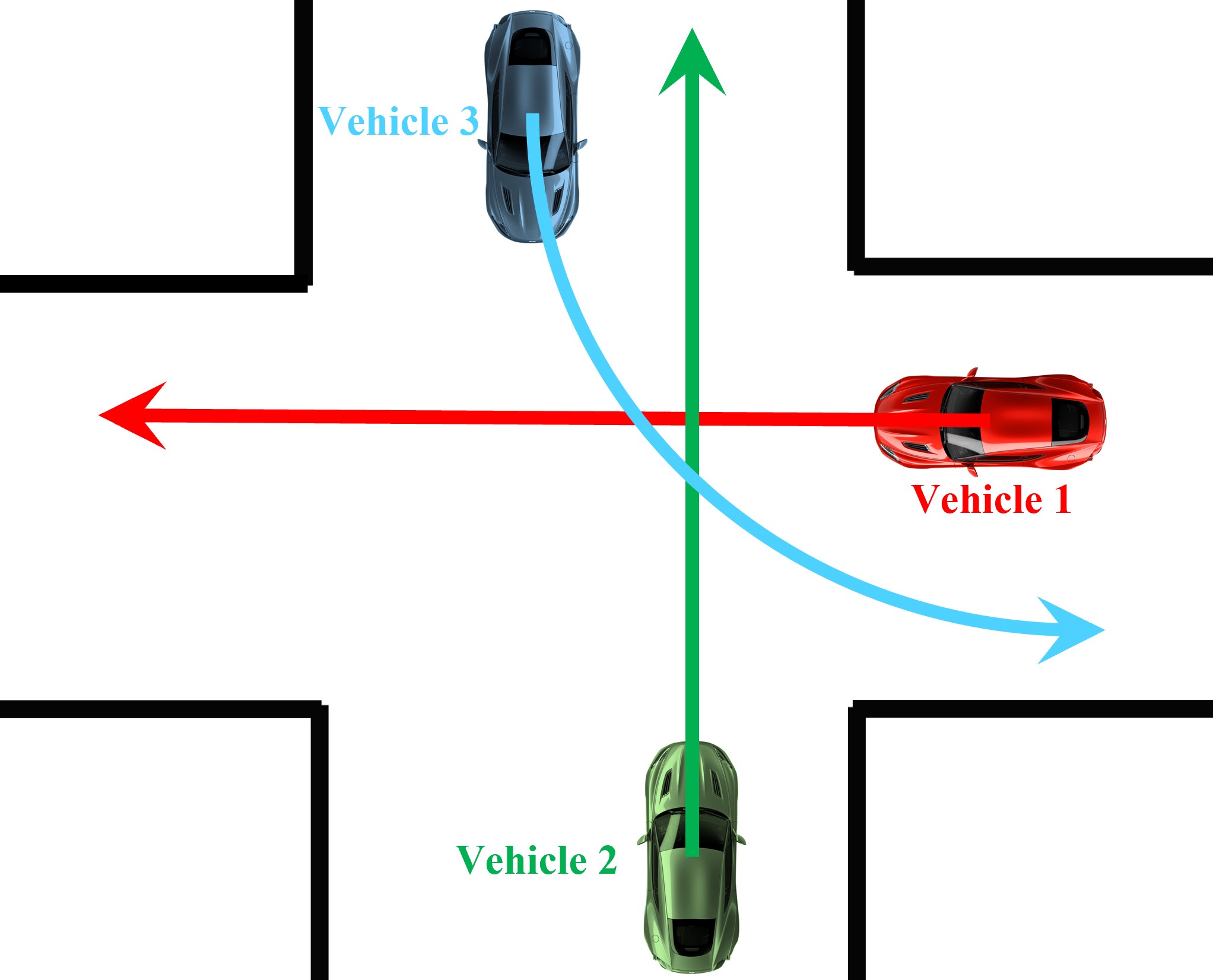}
	\caption{Three vehicles simulated at the unsignaled intersection.}
	\label{fig:intersection}
\end{figure}

\begin{figure*}[!t]
	\centering
	\vspace{-1.5em}
	\subfloat{
		\includegraphics[width=2in]{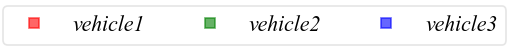}}
	\qquad\qquad\qquad\qquad\quad
	\subfloat{
	\includegraphics[width=2in]{./ttlegend.png}}
	\vspace{-1em}
	\\
	\subfloat[ISTCs results of group I]{
		\includegraphics[width=3in]{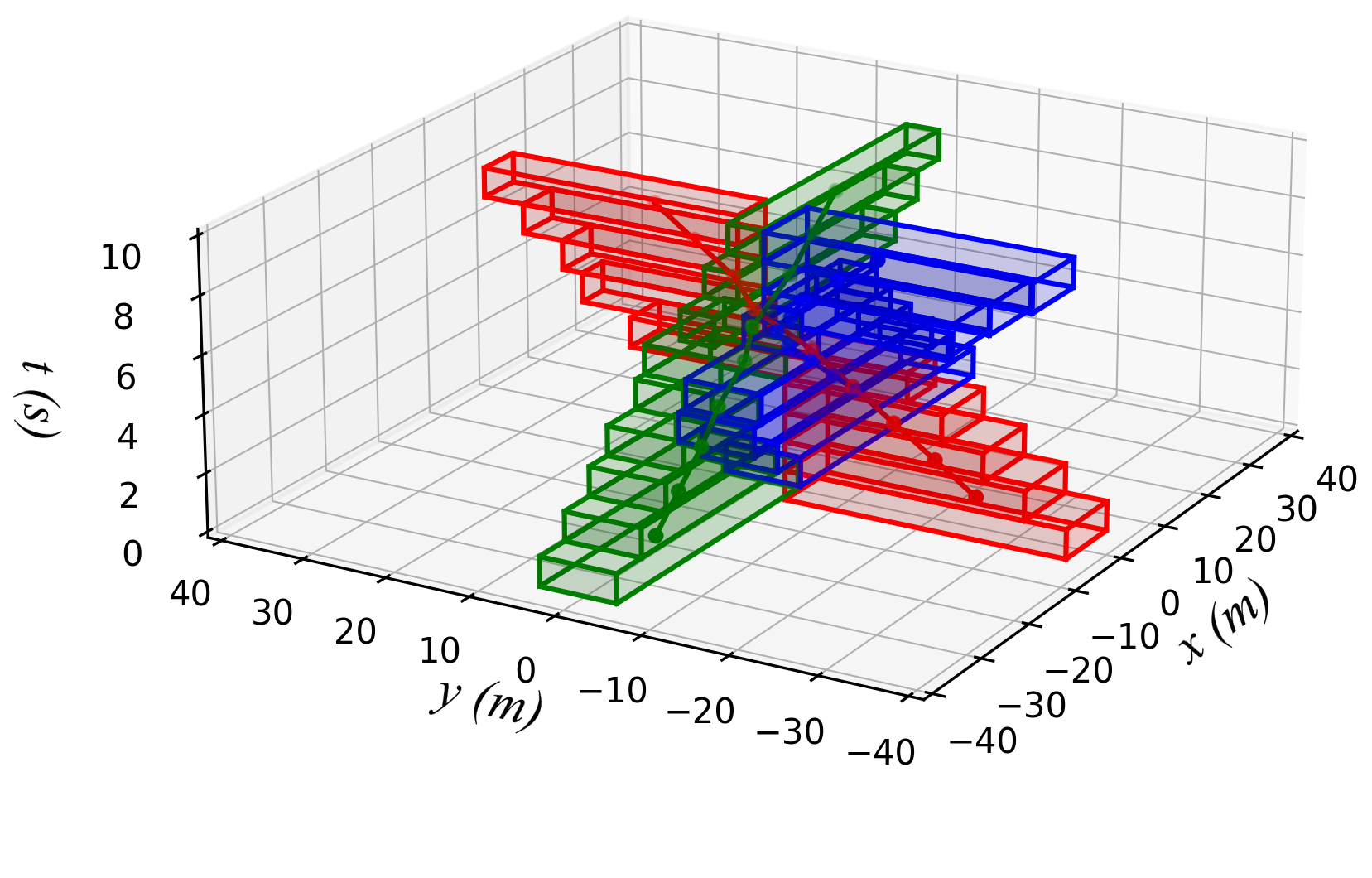}}
	\quad
	\subfloat[ISTCs results of group II]{
		\includegraphics[width=3in]{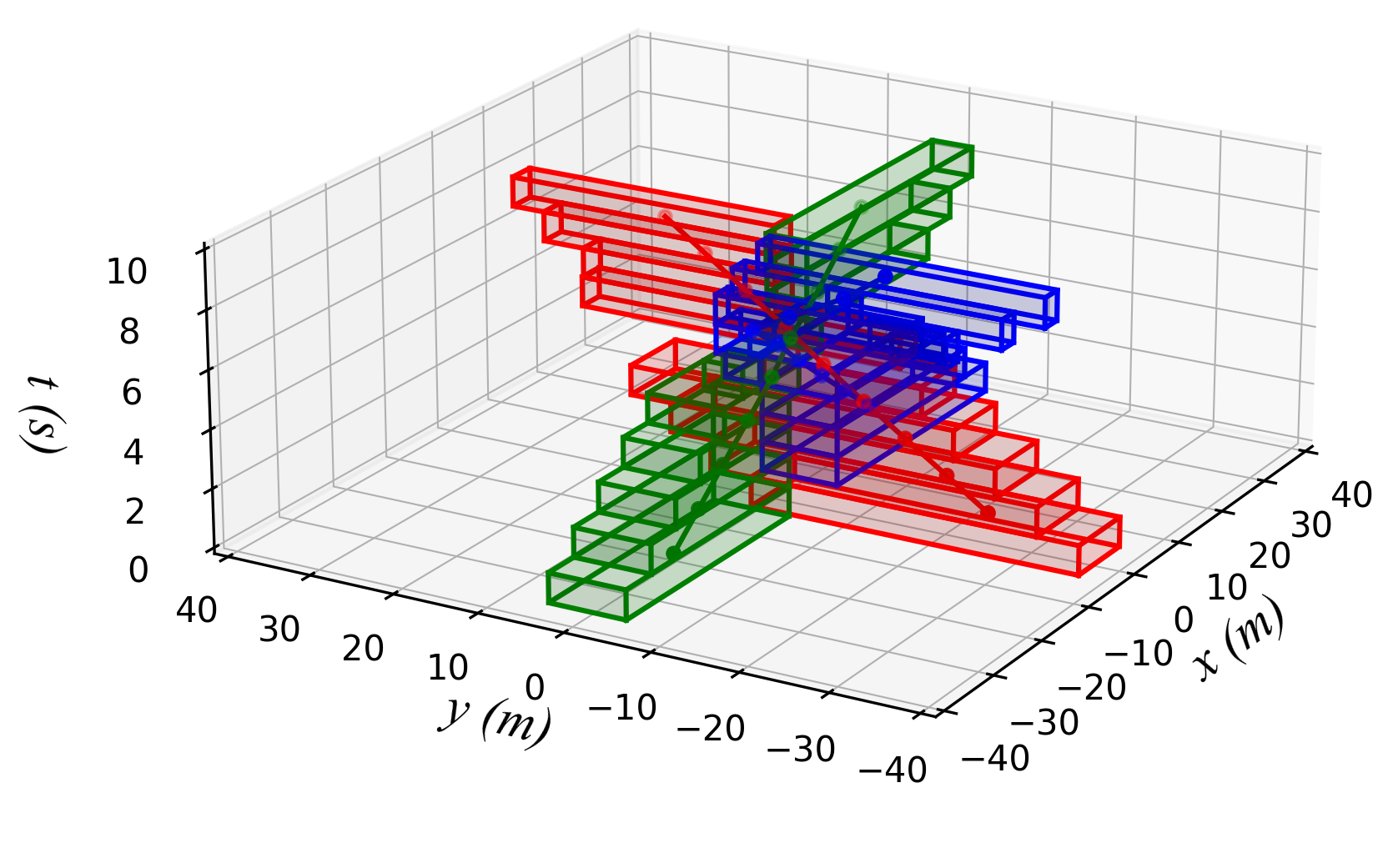}}
	\caption{Comparison results of the multi-vehicle ISTCs. (Inside each spatial temporal corridor are pivotal-points and their connecting lines)}
	\label{fig:Result1}
\end{figure*}
\begin{figure*}[!t]
	\centering
	\subfloat[vehicle $1$, ${\eta^1}=0.20$]{
		\includegraphics[width=1.8in]{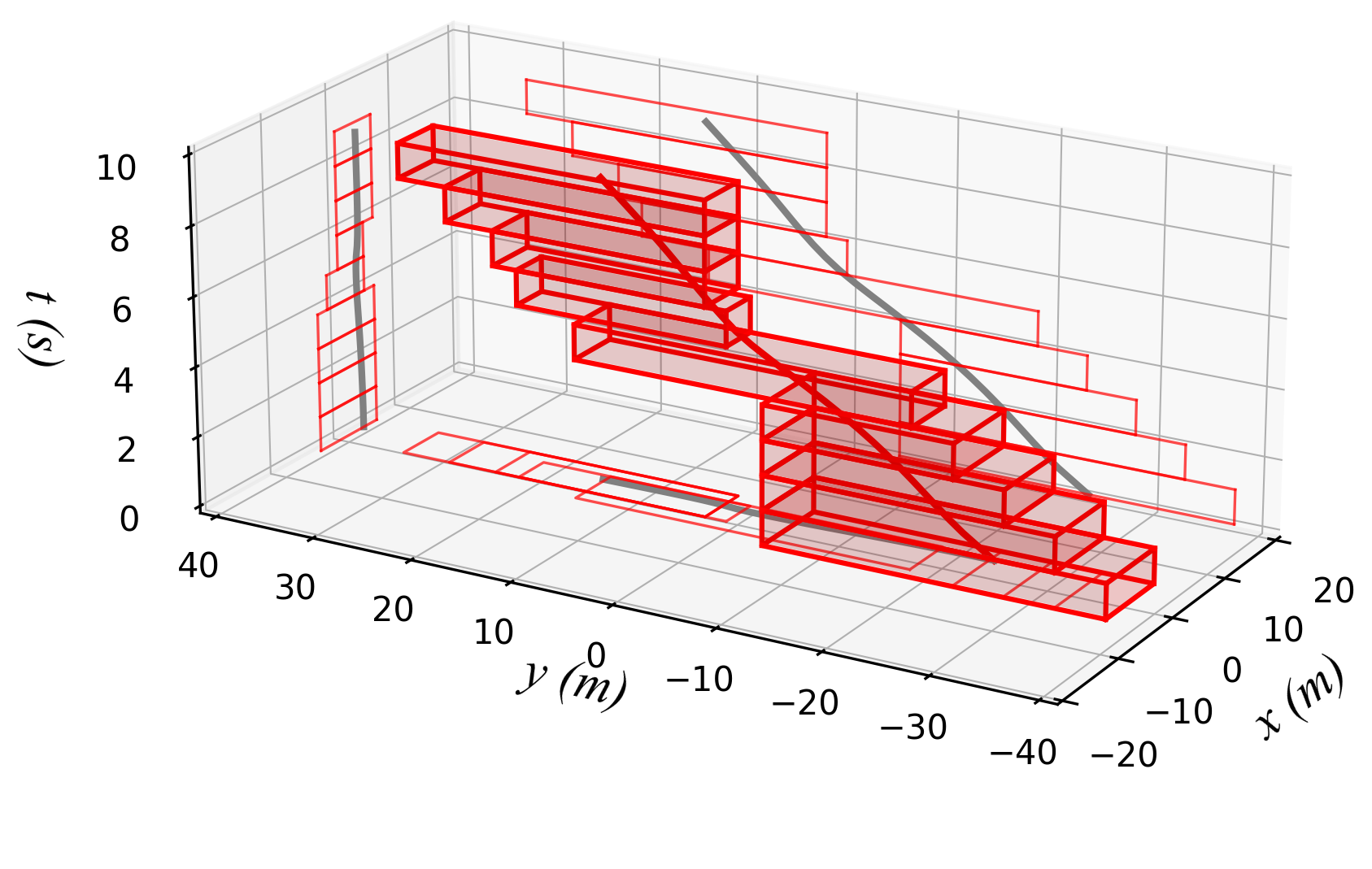}}
	\qquad
	\subfloat[vehicle $2$, ${\eta^2}=0.01$]{
		\includegraphics[width=1.6in]{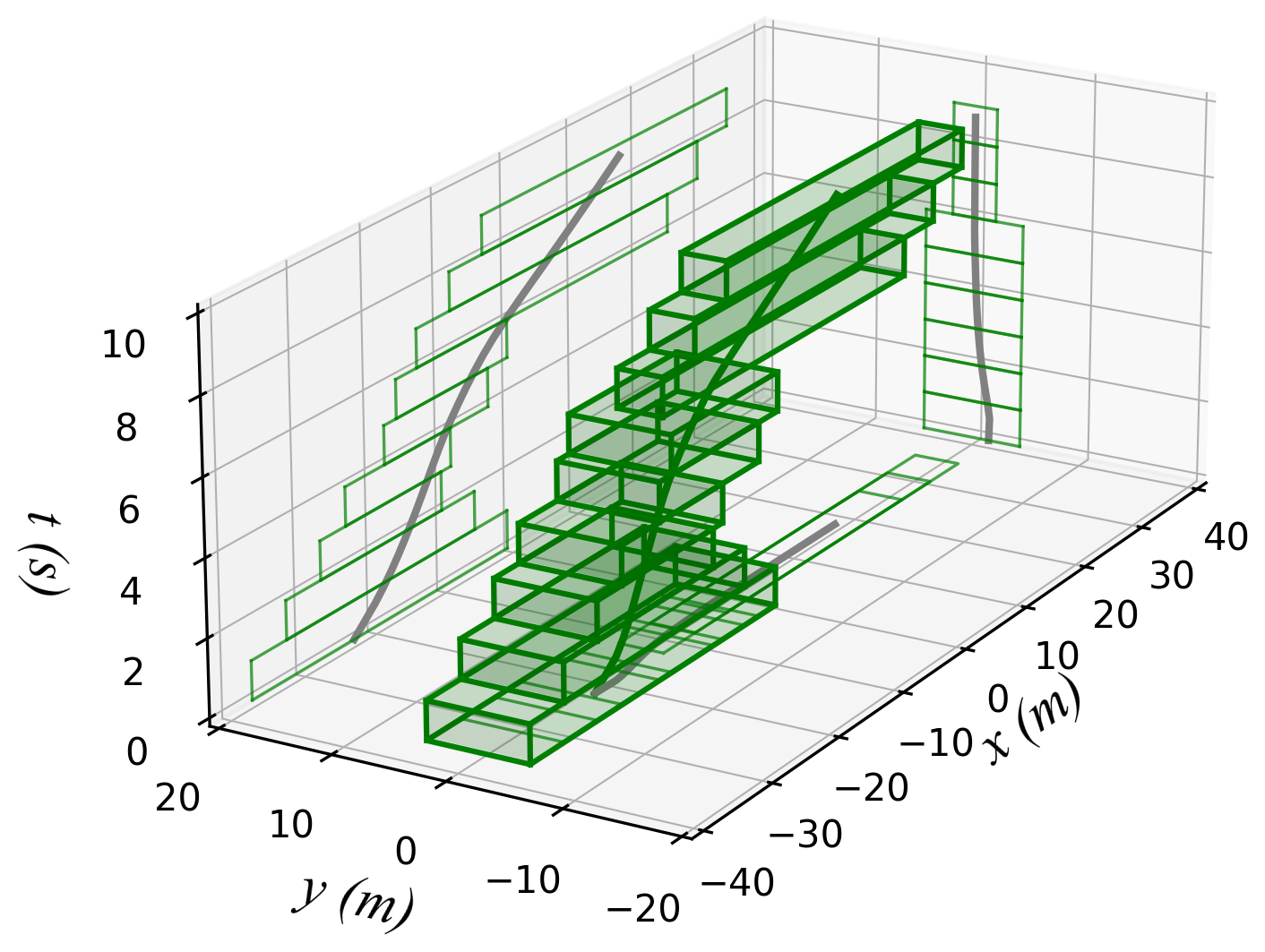}}
	\qquad
	\subfloat[vehicle $3$, ${\eta^3}=0.50$]{
		\includegraphics[width=1.8in]{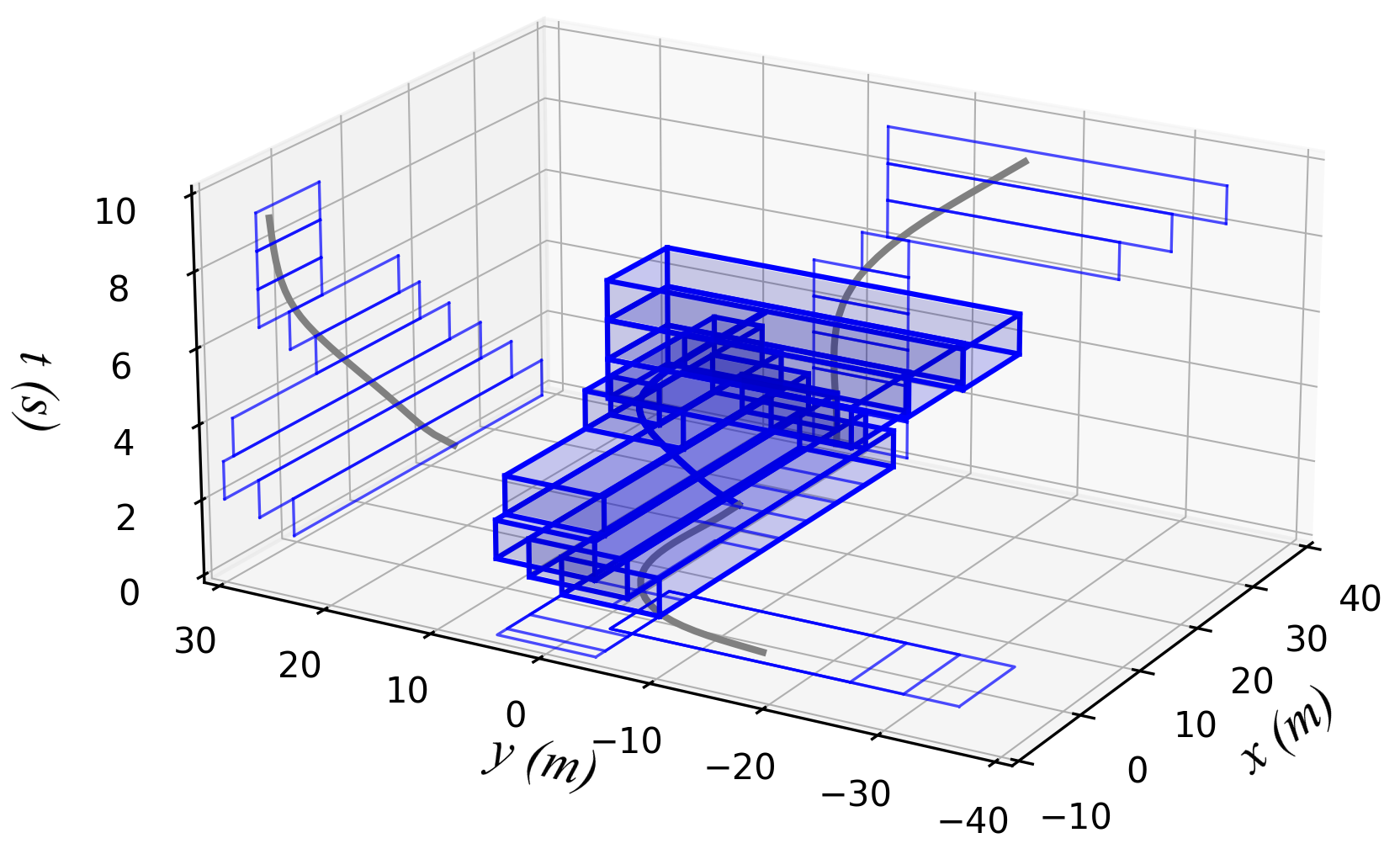}}
	\\
	\subfloat[vehicle $1$, ${\eta^1}=0.50$]{
		\includegraphics[width=1.8in]{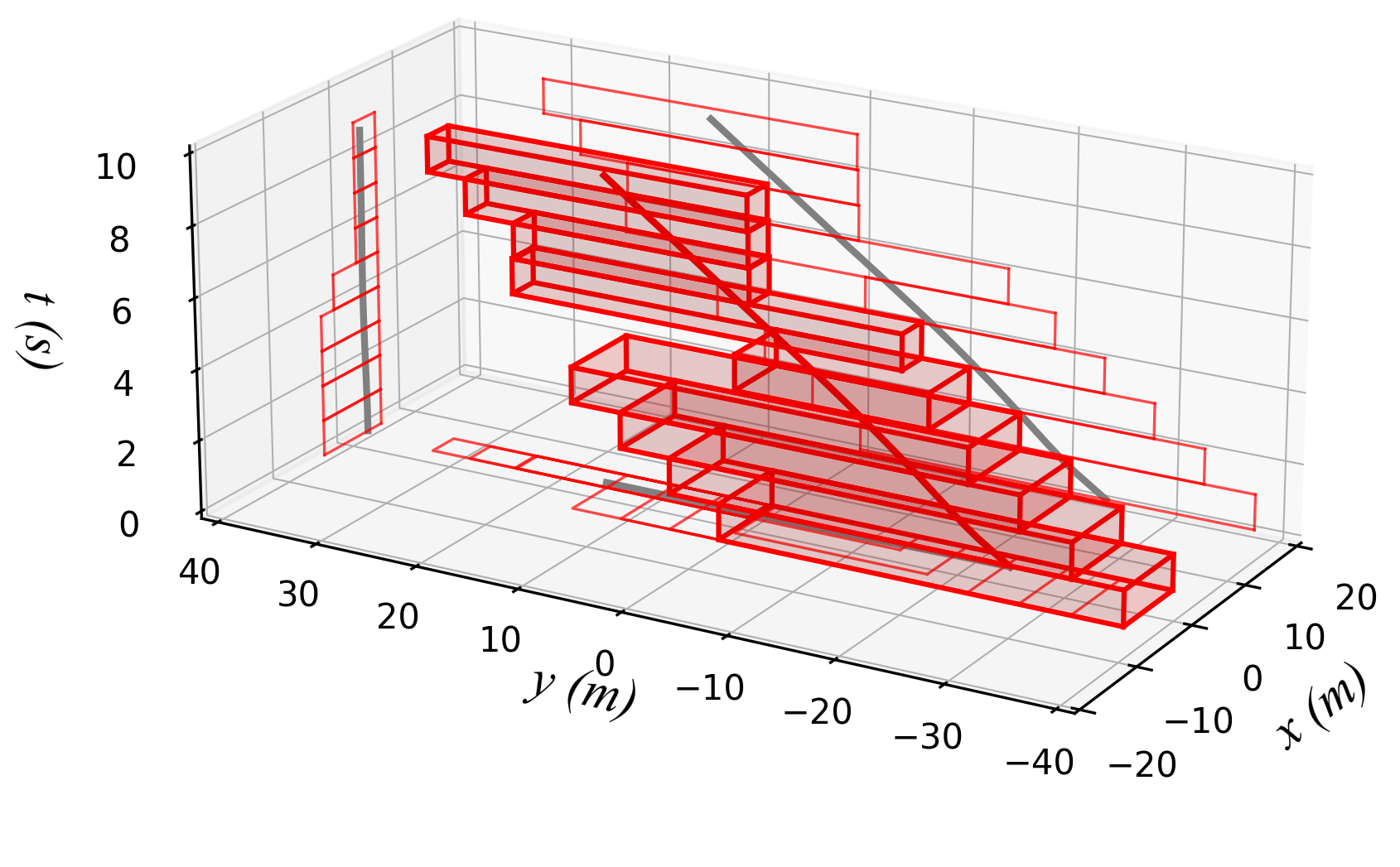}}
	\qquad
	\subfloat[vehicle $2$, ${\eta^2}=0.40$]{
		\includegraphics[width=1.6in]{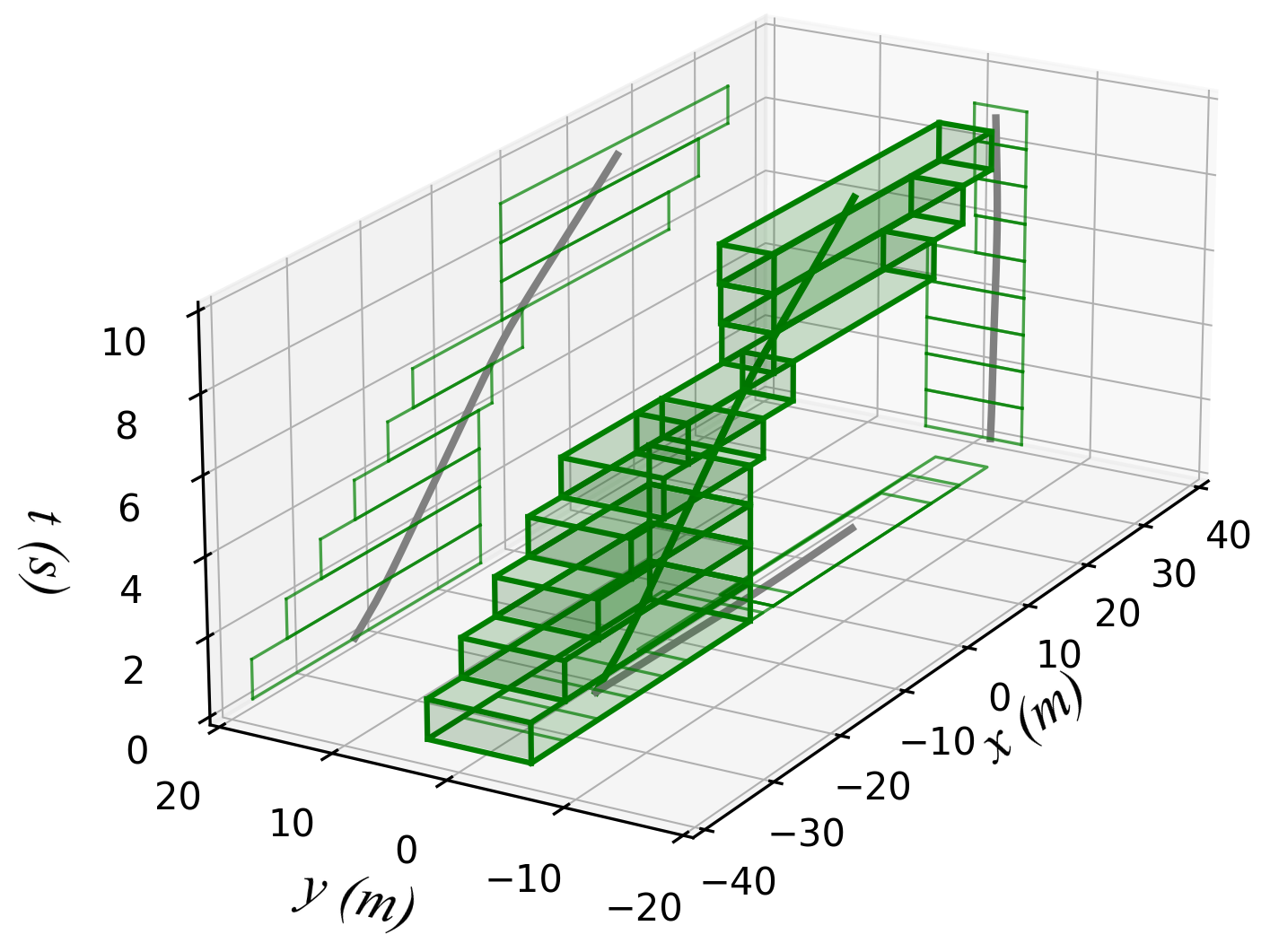}}
	\qquad
	\subfloat[vehicle $3$, ${\eta^3}=0.01$]{
		\includegraphics[width=1.8in]{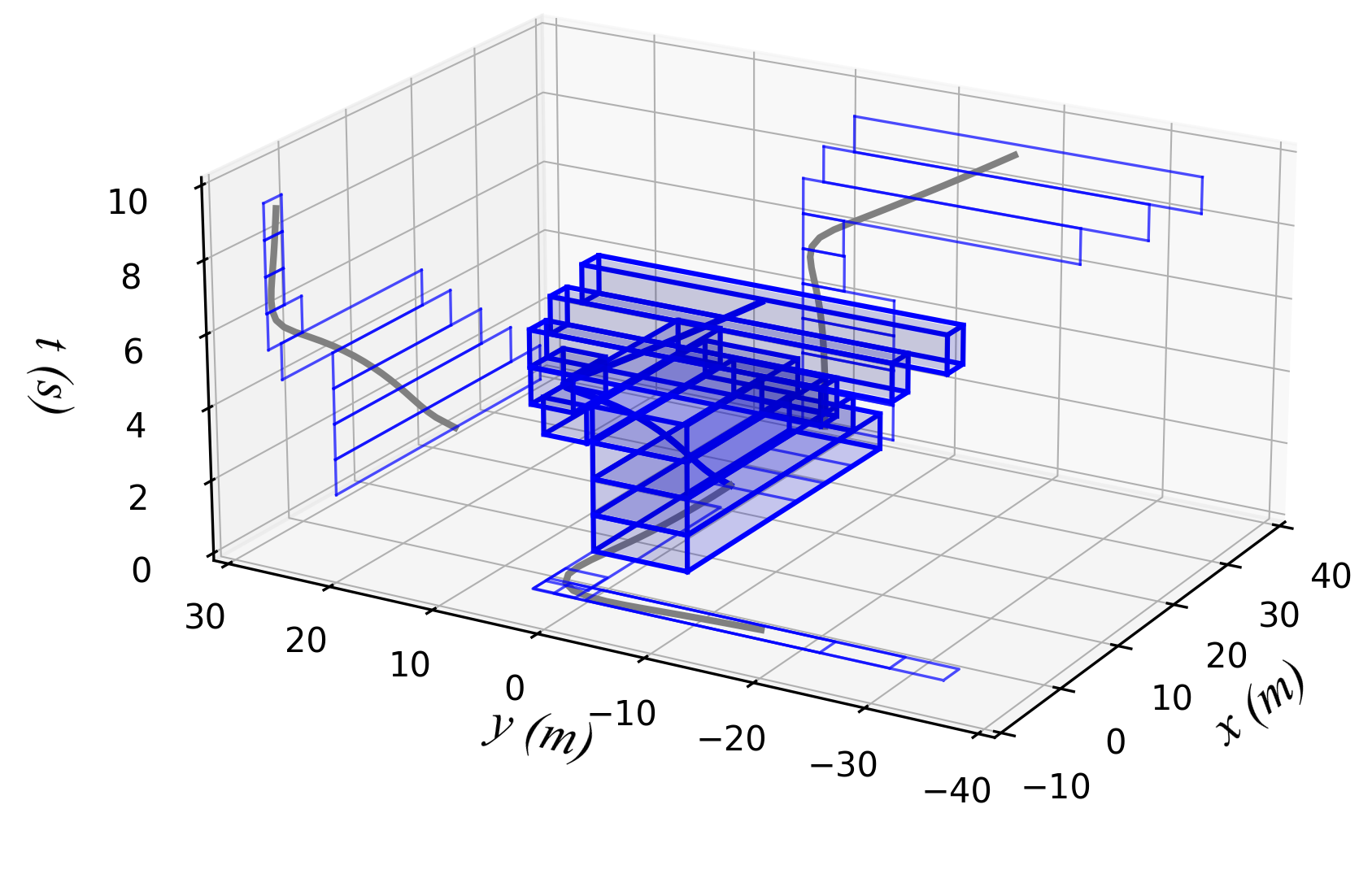}}
	\caption{ The results of the individual spatial temporal corridor with the final trajectory inside; (a)(b)(c) are the results of group I, and (d)(e)(f) are the results of group II.}
	\label{fig:Result2-1}
\end{figure*}
\vspace{-2em}

\subsection{Simulated experiments of  unsigenaled intersection}
\subsubsection{Description}
Simulation of three vehicles from different directions in an unsignalized intersection scenario is shown in Fig.\ref{fig:intersection}. $vehicle 1$ and $vehicle 2$ came from different directions with the intention of going straight, and $vehicle 3$ came from a different direction than the first two with the intention of turning left. In this structured scenario, the initial guidance trajectories can be projected based on the vehicles' intent in combination with the lane lines. All vehicles are assumed to pass at a uniform speed of $5m/s$. Afterwards, the initial guidance trajectory of each vehicle was indicated by different colored lines in Fig.\ref{fig:intersection}. It is assumed that the dimensional parameters of all vehicles are: length = 4$m$, width = 2$m$, wheelbase = 2.7$m$, and the width of the road is 10$m$. Besides, the relevant parameter settings of the algorithm are shown in Table~\ref{tab1}. Two sets of experiments are conducted as a comparison to reveal the effects of different priority weights. In the first set we let the left-turning $vehicle 3$ to have the higher priority, while in the second set we gave the higher priority to straight-going $vehicle 1$ and $vehicle 2$, as shown in Table~\ref{tab2}. In both the groups of experiments, the smallest priority weight was almost equal to $0$, in order to increase the gradient of objective function and increase the solving speed.

\begin{table}[!h]
	\renewcommand\arraystretch{1.5}
	\begin{center}
		\caption{Relevant Parameters of The Proposed Method}
		\label{tab1}
		\begin{tabular}{cc|cc|cc}
			\specialrule{0.1em}{0pt}{0pt}
			Parameters & Value & Parameters & Value & Parameters & Value\\
			\hline
			$\gamma{_{s+}^{1,2,3}}$ & 10 & $w{_{ref}^{1,2,3}}$ & 10.00 & $w{_j^{1,2,3}}$ & 1.0\\
			
			$\gamma{_{s-}^{1,2,3}}$ & 5 & $w{_{\mathcal{K}}^{1,2,3}}$ & 1.0 & $w{_{px}^{1,2,3}}$ & 1.0\\
			
			$w{_{area}^{1,2,3}}$ & 0.50  & $w{_{\beta}^{1,2,3}}$ & 100.0 & $w{_{py}^{1,2,3}}$ & 1.0\\
			\specialrule{0.1em}{0pt}{0pt}
		\end{tabular}
	\end{center}
\end{table}

\begin{table}[!h]
	\vspace{-2.0em}
	\begin{center}
		\caption{The Priority Weights of Each Vehicle in Turn-first Scenario and Straight-first Scenario}
		\label{tab2}
		\begin{tabular}{cccc}
			\specialrule{0.1em}{0.5pt}{1pt}
			\specialrule{0em}{0.5pt}{1pt}
			Group & ${\eta^1}$ & ${\eta^2}$ & ${\eta^3}$\\
			\hline
			I & 0.20 & 0.01 & 0.50\\
			II & 0.50 & 0.40 & 0.01\\
			\specialrule{0.1em}{0.5pt}{1pt}
		\end{tabular}
	\end{center}
\end{table}

\subsubsection{Results and discussion}
The constructed ISTCs of both groups can be generated as expected, shown in Fig.\ref{fig:Result1}. Based on the initial guidance trajectories, the corridor-cubes of each vehicle are expanded in the right direction until leaving the conflict zone. Besides, the spatial temporal corridors of different vehicles do not overlap in any time unit, ensuring safety during the conflict resolution. 

Comparing the results of ISTCs, the effect of different priority weights is distinct. Vehicle with higher priority weight tend to generate its spatial temporal corridor with larger area, and the pivotal-points of corridor-cubes tend to be more closer to the initial reference trajectory. Conversely, vehicle with lower priority weight is more inclined to sacrifice their driving possibilities to make space for other high-priority vehicles. But anyway, regardless of whether the priority weight is larger or smaller, the corridor of each vehicle is always safe and satisfying dynamics, due to the constraints settled in section IV. In group I, the spatial temporal corridor of $vehicle 3$ was found to be the most aggressive in occupying the center of the intersection at the beginning of the conflict. While in group II, the center of the intersection was mostly occupied by the corridor of $vehicle 1$. Furthermore, the results of the individual spatial temporal corridor with in-corridor trajectory are shown in Fig.\ref{fig:Result2-1}, which also expose the effect of priority weight. Comparing the corridor and trajectory of left-turning $vehicle 3$ shown in Fig.\ref{fig:Result2-1}(c) and Fig.\ref{fig:Result2-1}(f), the results of group I were significantly better than those of group II, mainly in two aspects: 1) The overall space occupied by corridor is greater. 2) The generated final trajectory is smoother. By contrast, when in the straight-first group with $\eta^3=0.01$, the final trajectory of $vehicle 3$ is highly restricted to the narrow corridor space. 

\begin{figure*}[!t]
	\centering
	\subfloat[Trajectory results of group I]{
		\includegraphics[width=3in]{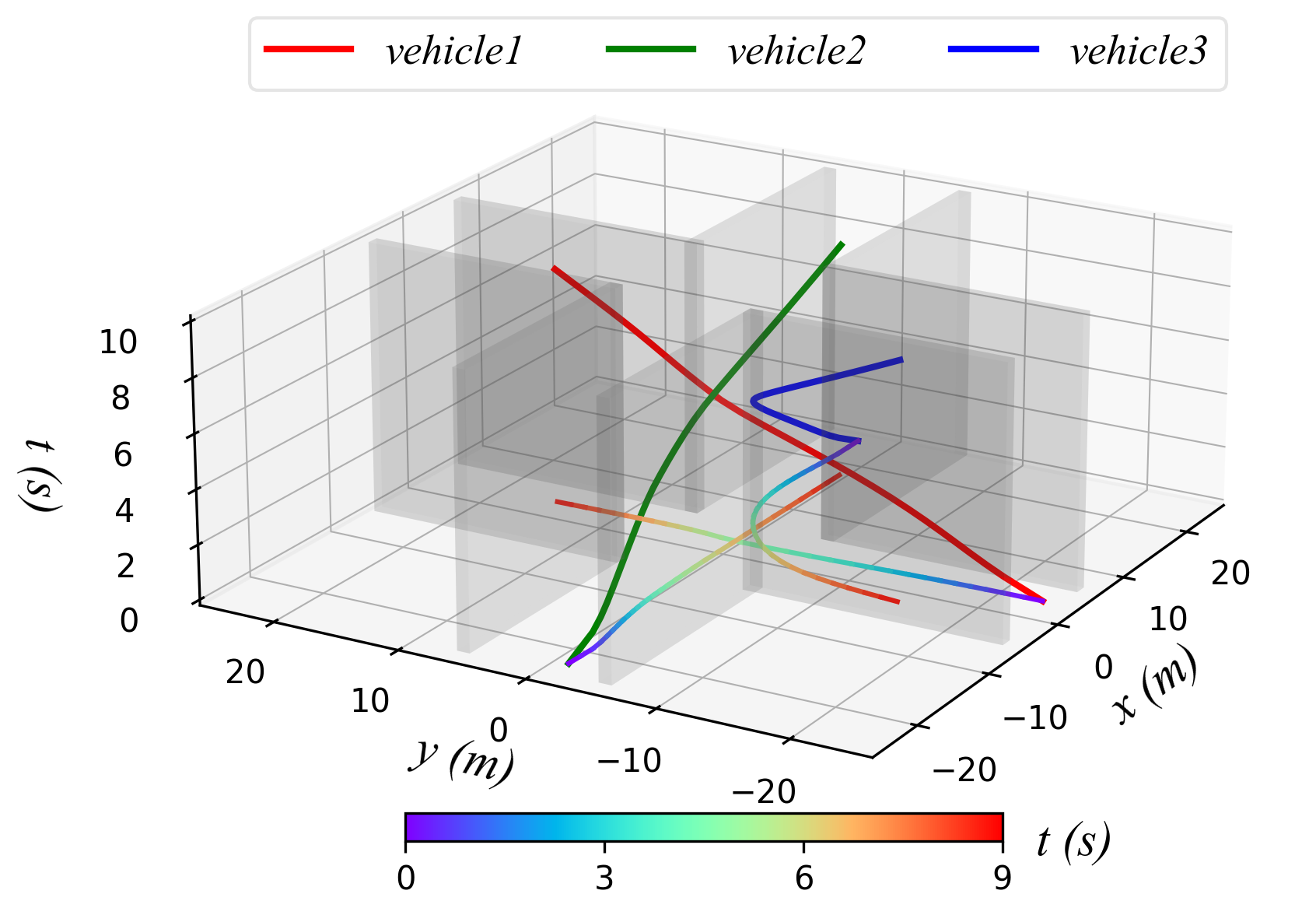}}
	\quad
	\subfloat[Trajectory results of group II]{
		\includegraphics[width=3in]{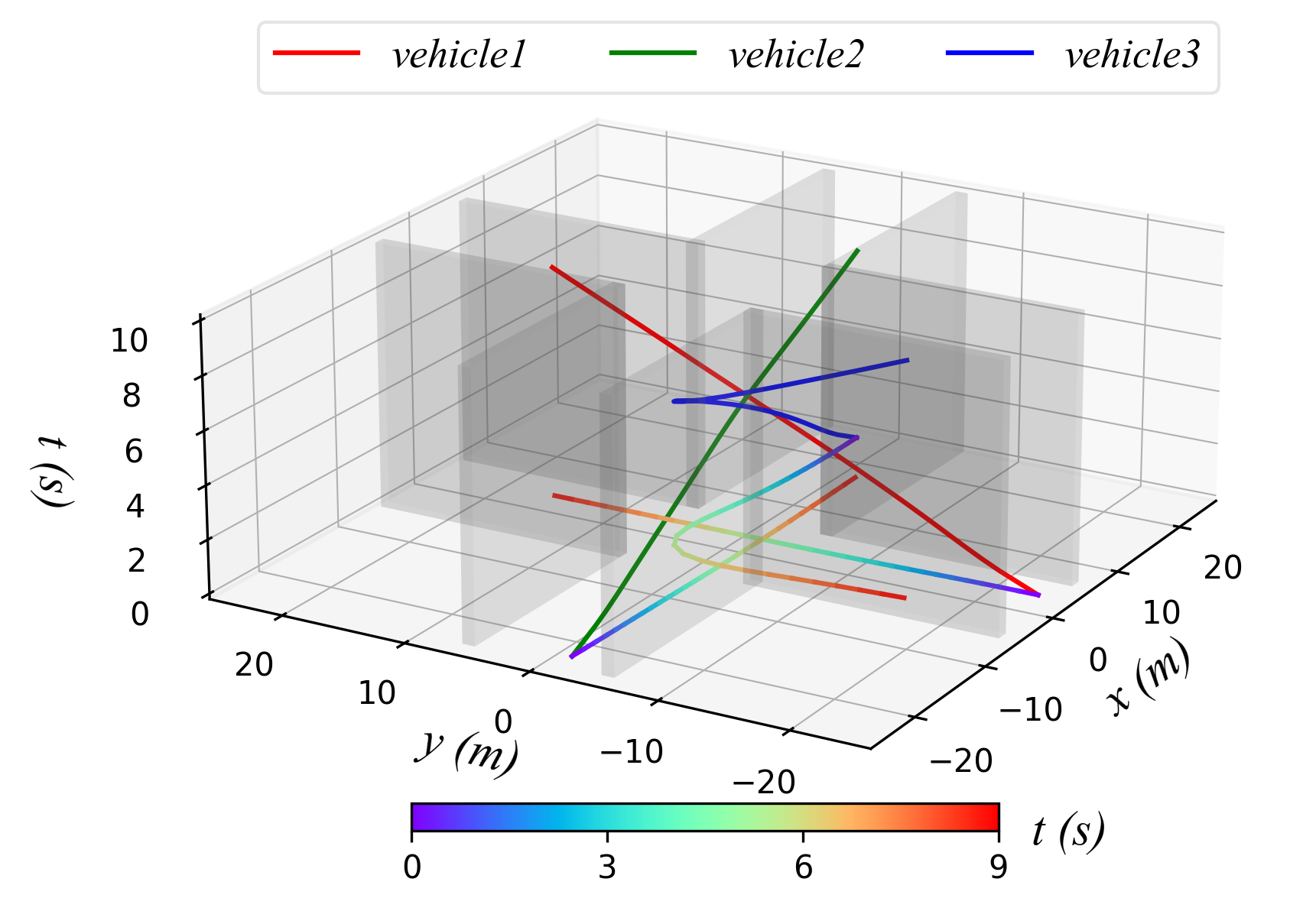}}
	\caption{Comparison results with the final multi-vehicle trajectories.}
	\label{fig:Result2-2}
\end{figure*}

\begin{figure}[!h]
	\centering
	\subfloat[Group I]{
		\includegraphics[width=1.7in]{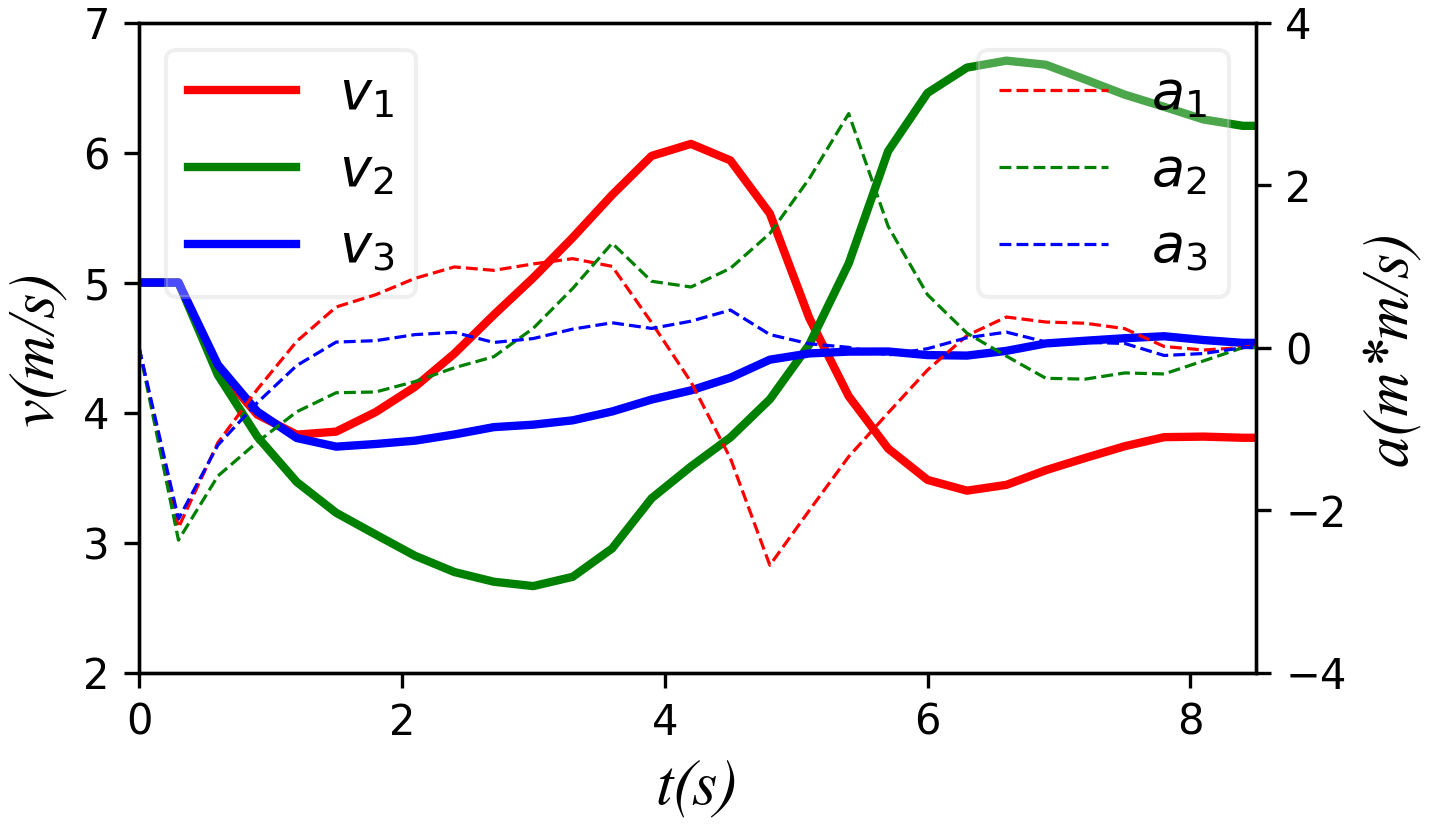}}
	\subfloat[Group II]{
		\includegraphics[width=1.7in]{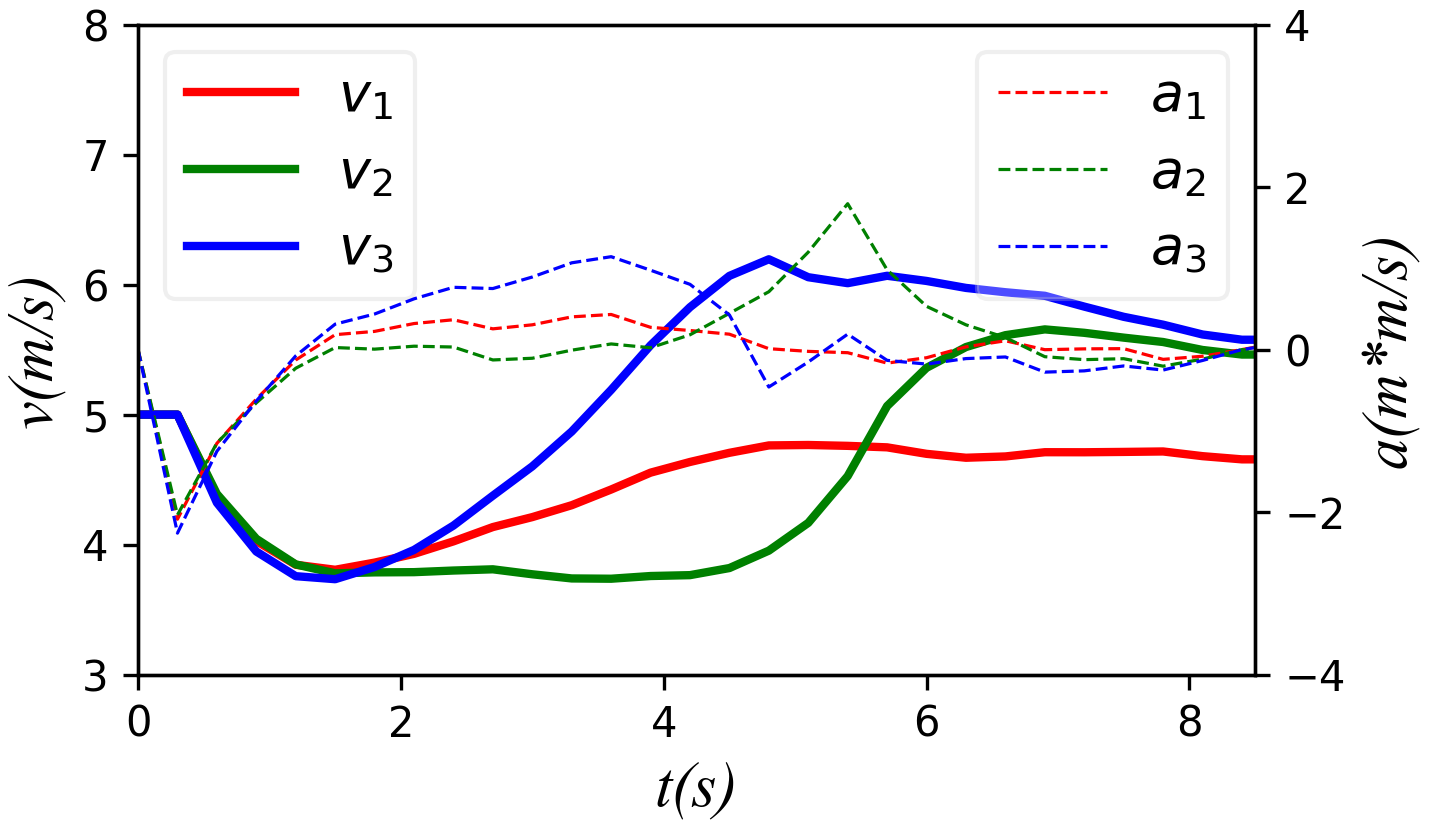}}
	\caption{ Velocity and acceleration results in unsignaled intersection}
	\label{fig:Result3}
\end{figure}

The multi-vehicle trajectory results as well as the velocity and acceleration results reflect the flexibility of the proposed method in the 3-D$(x,y,t)$ space-time configuration space ,which are respectively shown in Fig.\ref{fig:Result2-2} and Fig.\ref{fig:Result3}. In group I, $vehicle 3$ with the highest priority weight performs a comfortably large radius left-turn under initial guidance. Meanwhile $vehicle 1$ and $vehicle 2$ give way to $vehicle 3$ mainly by adjusting their respective speeds, according to Fig.\ref{fig:Result3}(a). By contrast, in group II, left-turning $vehicle 3$ chose to make a detour to the core conflict area and turned at a smaller radius with lower comfort. While $vehicle 1$ and $vehicle 2$ benefited from this and gained more stable speeds, as shown in Fig.\ref{fig:Result3}(a). Besides, it is demonstrated in both sets of experiments in Fig.\ref{fig:Result3} that all vehicles chose to slow down at the beginning of their involvement in the conflict. The magnitude of fluctuations in the speed and acceleration is negatively correlated with the relative magnitude of the vehicle's priority weight. 

The time consumption shown in Table~\ref{tab3} illustrates the efficiency of the proposed algorithm. Since the hierarchical framework is able to decouple multi-vehicle coordinated motion planning through ISTCs to individual vehicle for trajectory optimization, the time consumption can be viewed as two parts. In both experiment groups, the first layer of the framework was able to construct ISTCs rapidly. And in the second layer, each vehicle can obtain a safe and high-quality trajectory through the conflict area with an acceptable time consumption, fully considering of vehicle kinematic characteristics.

\begin{table}[!h]
	\renewcommand\arraystretch{1.2}
	\begin{center}
		\caption{The Time Consumption of The Whole Method}
		\label{tab3}
		\begin{tabular}{|c|c|c|c|c|c|}
			\hline
			\multirow{2}{*}{Group} & $t_{layer1}$ & \multicolumn{3}{c|}{$t_{layer2}$(s)}\\
			\cline{3-5}
			& (s) & $vehicle1$ & $vehicle2$ & $vehicle3$ \\
			\hline
			I & 0.092 & 0.146 & 0.042 & 0.167\\
			\hline
			II & 0.073 & 0.074 & 0.053 & 0.246\\
			\hline
		\end{tabular}
	\end{center}
\end{table}

To sum up, the experimental results in unsignalized intersection are shown to be consistent with the original intention of the method detailed in Section IV. ISTCs can safely and efficiently resolve interaction conflicts and simplify the problem of generating collaborative trajectories. Flexible parameter settings allow vehicles to adopt different strategies when passing through conflict zones under the same environment and initial conditions. Besides, finding the solution in the 3D spatio-temporal configuration space greatly increases the flexibility of the vehicle to avoid conflicts.

\subsection{Comparison with Baseline Algorithms in the Challenging Dense Scenario}
\subsubsection{Description}
In order to verify the feasibility and advantage of proposed hierarchical method in general scenarios, we conducted comparison experiments with CL-CBS proposed in \cite{wen2022cl}. The experiments were carried out in a 50m*38m map in which 4 vehicles, 6 vehicles and 8 vehicles were tested with and without obstacles in the middle of the map. In order to increase the challenge of resolving conflicts, the initial position of vehicles were set at the left or right sides of the map, and all of them were required to move to the other side without collision. In our method, the initial guide trajectories are found using Hybrid A*, and a constant velocity of 5m/s is set. In the simulation of both methods, the start and goal states are set to be the same. 

\begin{figure*}[!t]
	\centering
	{\includegraphics[width=6in]{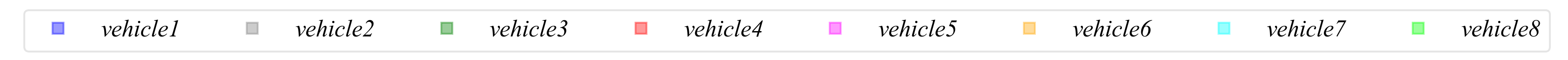}\vspace{-1em}}
	\subfloat[CL-CBS, obstacle-free]{
		\includegraphics[width=2.77in]{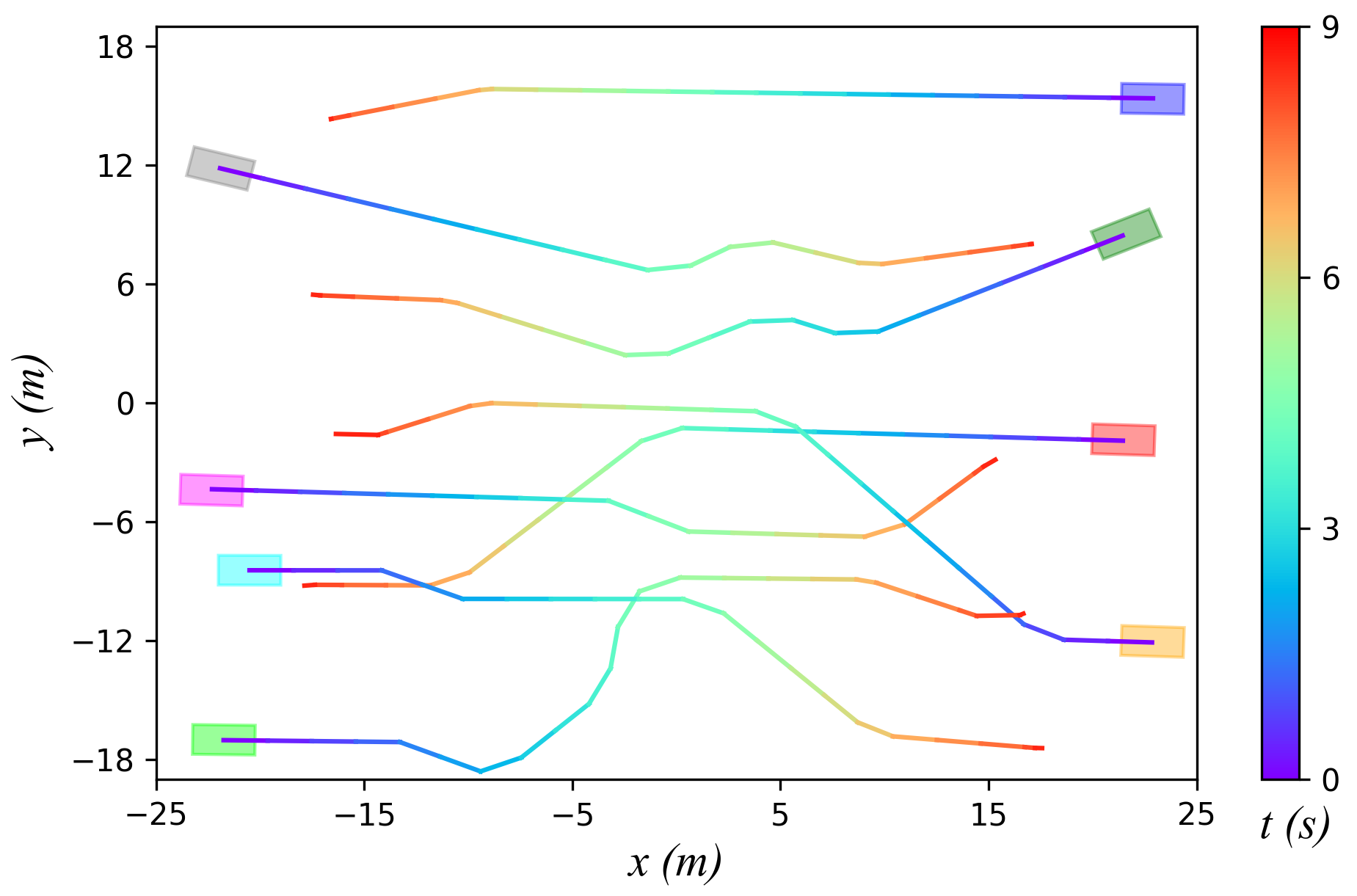}}
	\qquad
	\subfloat[Proposed Method, obstacle-free]{
		\includegraphics[width=2.77in]{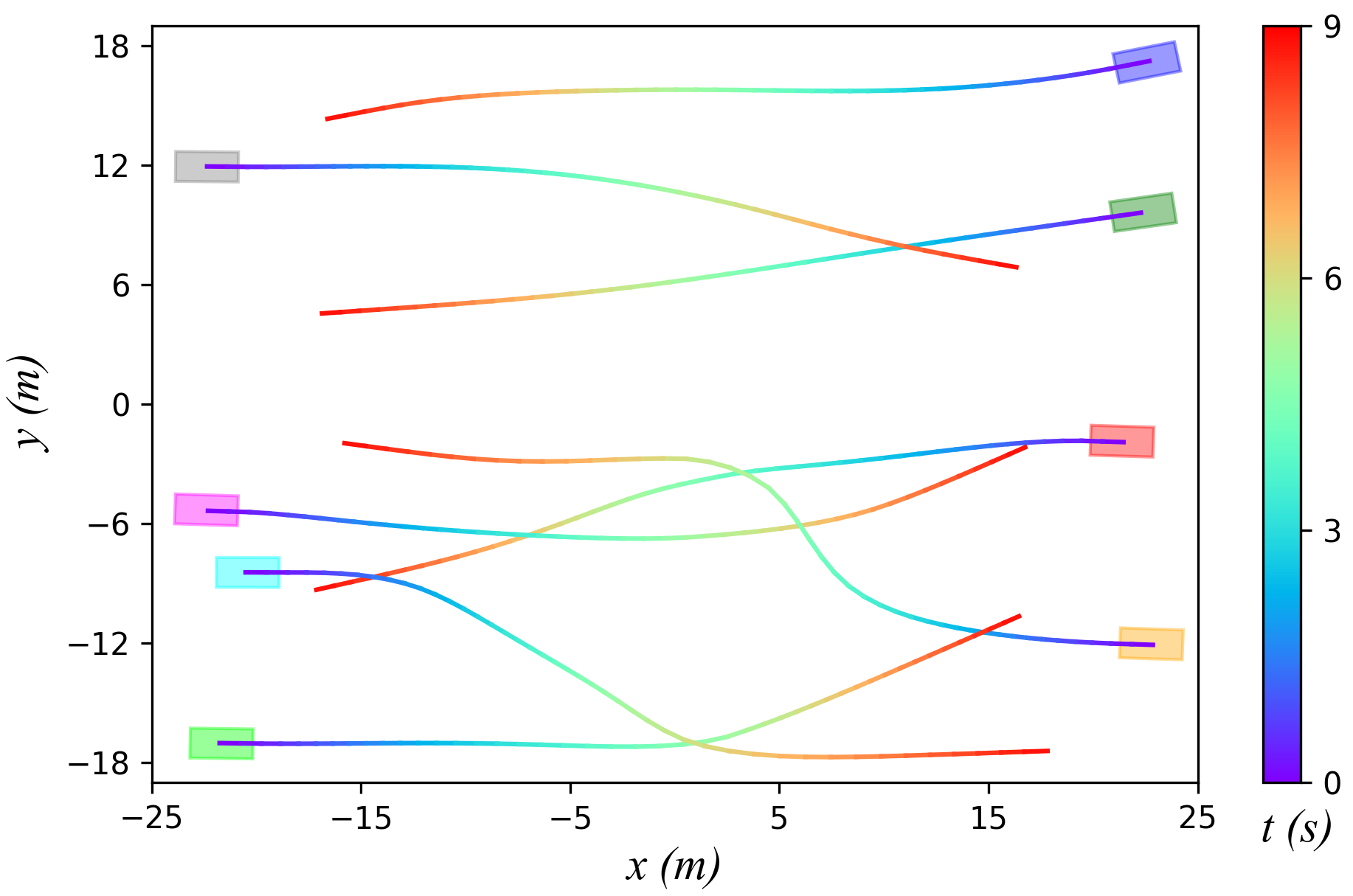}}
	\\
	\subfloat[CL-CBS, with obstacles]{
		\includegraphics[width=2.77in]{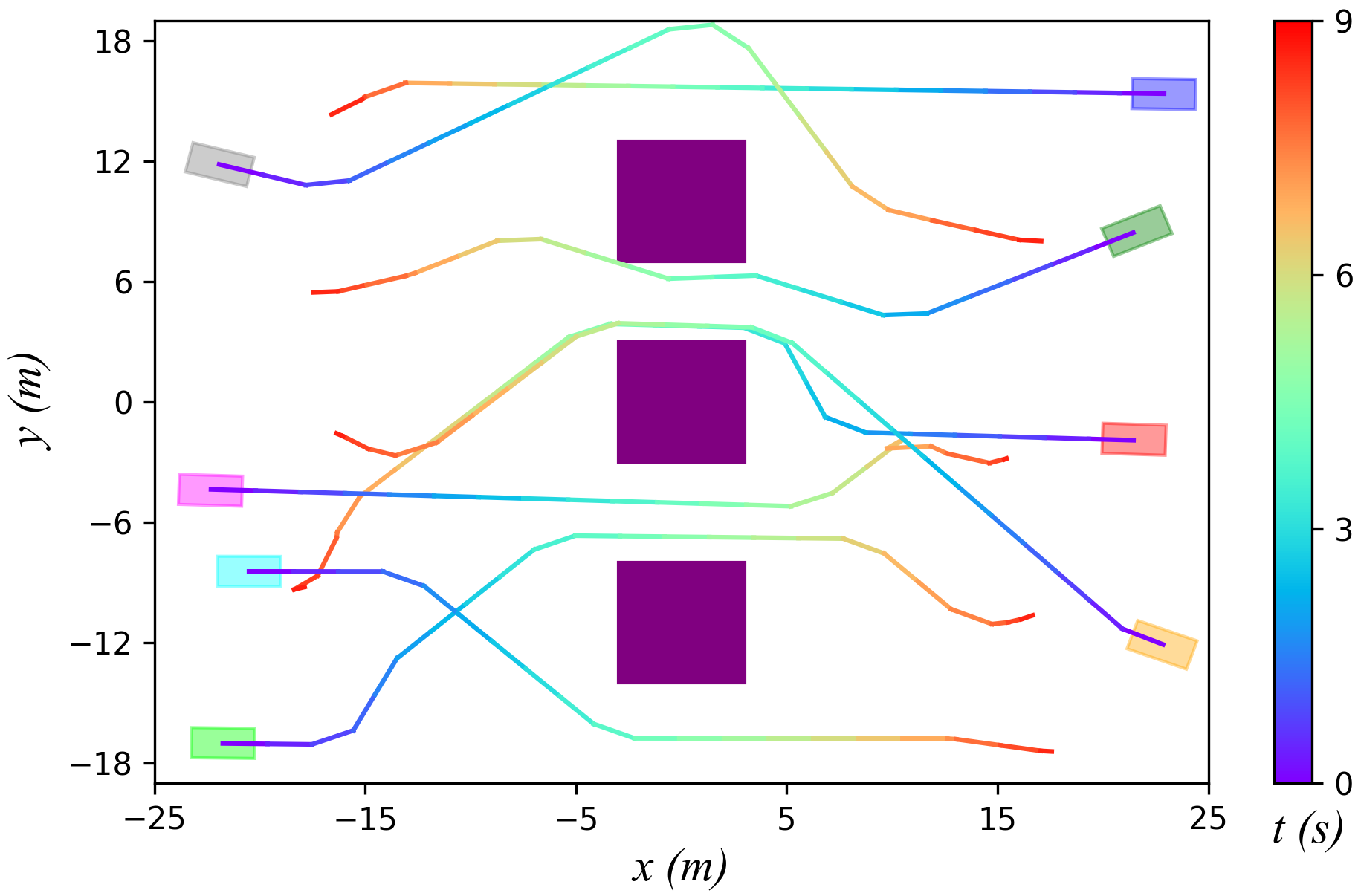}}
	\qquad
	\subfloat[Proposed Method, with obstacles]{
		\includegraphics[width=2.77in]{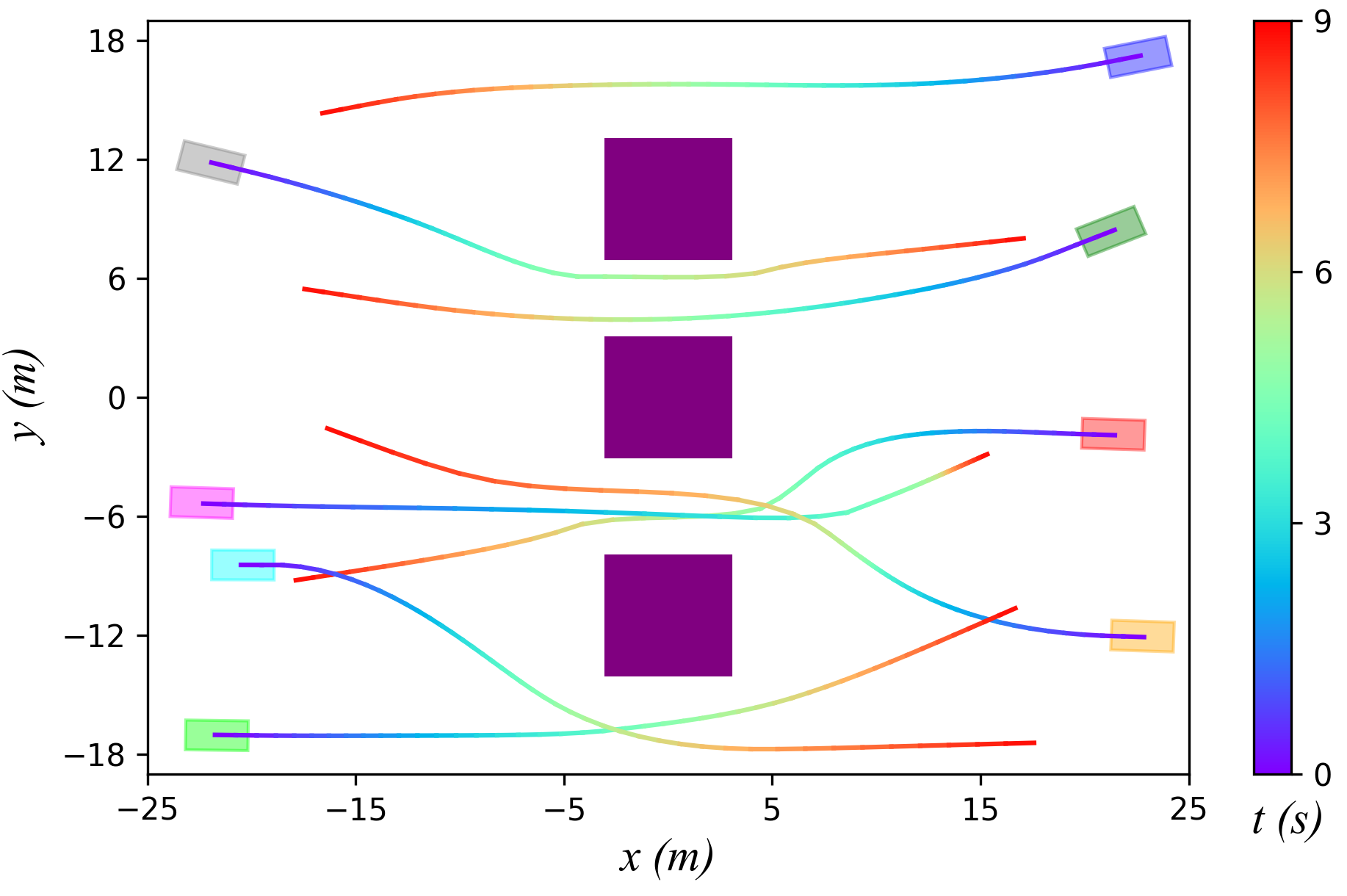}}
	\caption{ Comparison results with Baseline Algorithms in the Challenging Dense Scenario}
	\label{fig:Result4}
\end{figure*}

The comparison results are shown in Table~\ref{tab4}.  For CL-CBS, $N_{node}$ is the number of conflict nodes that have been expended by conflict-based-search, $t_{total}$ is the total time consumption of accomplish the multi-vehicle planning task, $L_{total}$ is the total length of the trajectories of all vehicles. For proposed method, $t_{l_1}$ is the time consumption of ISTCs construction in the first layer, $t_{l_2}$ is the time consumption of trajectories optimization for all the vehicles in the second layer, $t_{total}$ and $L_{total}$ have the same meaning as in CL-CBS. Besides, the trajectories of 8 vehicles driving with and without obstacles are listed in Fig.\ref{fig:Result4}.

\begin{table*}[!t]
	\renewcommand\arraystretch{1.1}
	\begin{center}
		\caption{Comparison Results with Baseline Algorithms in the Challenging Dense Scenario.}
		\label{tab4}
		\begin{tabular}{|c|c|c|c|c|c|c|c|c|}
			\hline
			Vehicle & \multirow{2}{*}{Obstacles} & \multicolumn{3}{c|}{CL-CBS} & \multicolumn{4}{c|}{Proposed Method} \\
			\cline{3-9}
			numbers & & $N_{node}$ & $t_{total}(s)$ & $L_{total}(m)$ & $t_{l_1}(s)$ & $t_{l_2}(s)$ & $t_{total}(s)$ & $L_{total}(m)$ \\
			\hline
			4 & \usym{2613} & 5 & 0.069 & 175.554 & 0.171 & 0.402 & 0.573 & 157.615\\
			\hline
			4 & \usym{1F5F8} & 3 & 0.065 & 182.119 & 0.215 & 0.415 & 0.630 & 159.735\\
			\hline
			6 & \usym{2613} & 6 & 0.228 & 258.018 & 0.651 & 0.793 & 1.444 & 235.376\\
			\hline
			6 & \usym{1F5F8} & 225 & 9.611 & 272.611 & 1.622 & 0.774 & 2.436 & 239.282 \\
			\hline
			8 & \usym{2613} & 530 & 22.98 & 352.066 & 3.954 & 0.977 & 4.931 & 313.634 \\
			\hline
			8 & \usym{1F5F8} & 737 & 38.090 & 381.616 & 6.369 & 1.002 & 7.371 & 319.149 \\
			\hline
		\end{tabular}
	\end{center}
\end{table*}

\subsubsection{Results and discussion}
The time consumption in both methods generally increases with the presence of obstacles and the number of vehicles, but our method is less sensitive to these factors and shows better performance in denser scenes. In proposed framework, the first layer of MIQP was able to construct ISTCs for all participating vehicles efficiently in all simulations. Benefit from ISTCs decoupling the multi-vehicle collaboration problem, the computational complexity of the second layer for each individual vehicle is not affected by the increasing complexity of the whole problem, so that the total time consumption for the second layer is positively linearly related to the number of vehicles. In contrast, CL-CBS saw an explosive increase in the number of conflict nodes in denser scenes that full of conflicts, resulting in a sharp escalation in time consumption. The reason for this phenomenon is that CL-CBS does not directly consider the collaboration between vehicles in the low-level planner, but first plans trajectories for each vehicle independently and then extends the conflict nodes by detecting conflicting positions. For each conflict node, CL-CBS also does not coordinate all the conflict-involved vehicles at the same time to resolve the conflict, but only applies a constraint to one vehicle at a time to make it bypass the conflict location.

Besides, in all the simulations, our method allowed vehicles to pass through the conflict zone with a relatively shorter path, because the ISTCs construction and trajectory optimization can adjust both the position and speed of vehicles in the 3-D $(x,y,t)$ configuration space, preserving a larger solution space. The results in Fig.\ref{fig:Result4} also demonstrate that our method can generate more delicate and smooth trajectories. As shown in Fig.\ref{fig:Result4}(d), it can be found that ISTCs can make full use of the lager solution space to improve passage efficiency while ensuring vehicle safety. For example, the trajectories of $vehicle 2$ and $vehicle 3$ met in the narrow opening and passed through in parallel, while the trajectory of $vehicle 5$ quickly passed through the nearest channel to make space for $vehicle 4$ and $vehicle 6$. By contrast in Fig.\ref{fig:Result4}(c), the trajectories generated by CL-CBS was over-discretized, even leading to cross the border or collide with obstacle, such as the trajectories of $vehicle 2$ and $vehicle3$.

However, CL-CBS based on conflict-based-search is extremely efficient in finding feasible trajectories in the case of sufficient free space, much faster than the ISTCs-based method proposed in this paper. The reason for this phenomenon is that the conflict-resolution mechanism of CL-CBS is to iteratively replan the trajectory of a single conflicting vehicle under the spatio-temporal constrain. In scenes with fewer vehicles and more free space, this mechanism has a low probability of causing new conflicts after resolving the current conflict, which greatly reduces the times of conflict nodes expansion and hybrid A* replannning. Differently, our method is to utilize a bi-level optimization of MIQP and NLP to globally coordinate all the vehicles simultaneously, so that takes longer to generate smoother trajectories.

\section{Conclusion}
In this study, a novel hierarchical framework is proposed to accomplish multi-vehicle coordinated motion planning in general scenes. MIQP designed in the first layer can efficiently construct ISTCs, opening up safe spatial temporal corridors for each vehicle in complex interaction environments. The constraints imposed in the MIQP ensure the safety of vehicles while preserving enough margin for the feasible trajectory generation. Based on this, NLP settled in the second layer is capable of effectively generating safe and smooth in-corridor trajectories with the consideration of non-holonomic kinematics. Benefiting from the whole framework is optimized in the 3D $(x,y,t)$ space-time configuration space, a larger solution space is preserved and the quality of the planning results is improved. Compared to other existing methods, the proposed bi-level method is able to decouple the global multi-vehicle motion planning task to individuals, simplifying the complexity and dispersing the computational pressure.

In future work, we will try to design more flexible spatial temporal corridor structures than cubes to further improve the adaptability and robustness of the algorithm.

\bibliographystyle{IEEEtran}
\bibliography{ref}

\begin{IEEEbiographynophoto}
	{Xiang Zhang} 
	received his B.S degree in 2022 in vehicle engineering from the Beijing Institute of Technology, Beijing, China, where he is currently pursuing the M.S. degree with the Intelligent Vehicle Research Center, School of Mechanical Engineering, Beijing Institute of Techonology.
	
	His research interests include intelligent vehicles, motion planning and control.
\end{IEEEbiographynophoto}

\begin{IEEEbiographynophoto}
	{Boyang Wang} 
	received his B.S degree in vehicle engineering from the Beijing Institute of Technology, Beijing, China, in 2013, and the Ph.D. degree from the Beijing Institute of Technology, Beijing, China, in 2020. He was a visiting researcher with
	the Interaction Digital Human Group of CNRS-UM LIRMM from 2017-2019. He was a Postdoctoral researcher with the Key Laboratory of Machine Perception(MOE), Peking University from 2020-2022. He is currently an assistant professor at the School of Mechanical Engineering, Beijing Institute of Technology.
	
	His interests include intelligent vehicles, driver behavior, motion planning and control, vehicle dynamics modeling.
\end{IEEEbiographynophoto}

\begin{IEEEbiographynophoto}
	{Yaomin Lu} 
	received her B.S degree in vehicle engineering from the Beijing Institute of Technology, Beijing, China, in 2019 and degree from the Beijing Institute of Technology, Beijing, China, in 2022. She now works in Baidu Intelligent Driving Group, Beijing, China.
	
	Her research interests include intelligent vehicles, motion planning and optimization method.
\end{IEEEbiographynophoto}

\begin{IEEEbiographynophoto}
	{Haiou Liu} 
	 received the B.S. and Ph.D. degrees from the Beijing Institute of Technology, Beijing, China, in 1998 and 2003, respectively. She was a Visiting Scholar with the Energy and Automotive Research Laboratory, Mechanical Engineering Department, Michigan State University, East Lansing, MI, USA, from September 2013 to September 2014. She is currently a Professor with the School of Mechanical Engineering, Beijing Institute of Technology. Her teaching interests focus on vehicle control classes at both undergraduate and graduate levels. Her current research interests include design and control of automated manual transmission and hybrid powertrain.
\end{IEEEbiographynophoto}

\begin{IEEEbiographynophoto}
	{Jianwei Gong}
	received his B.S. degree from the National University of Defense Technology, Changsha, China, in 1992, and the Ph.D. degree from Beijing Institute of Technology, Beijing, China, in 2002. He was a visiting researcher of Robotic Mobility Group, Massachusetts Institute of Technology, between 2011 and 2012. He is currently a professor and director of the Intelligent Vehicle Research Center, School of Mechanical Engineering, Beijing Institute of Technology.
	
	His interests include intelligent vehicle environment perception and understanding, driver behavior, motion planning and control.
\end{IEEEbiographynophoto}

\begin{IEEEbiographynophoto}
	{Huiyan Chen} 
	received the Ph.D. degree from the Beijing Institute of Technology, Beijing, China, in 2004. He has been working at Beijing Institute of Technology since 1981 and has served as the director of the Intelligent Vehicle Research Center. He is now a professor at the School of Mechanical Engineering, Beijing Institute of Technology.
	
	His interests include intelligent vehicles, powertrain system modeling an control, information technologies.
\end{IEEEbiographynophoto}

\vfill

\end{document}